\documentclass[letterpaper]{article} 
\usepackage{aaai2026}  
\usepackage{times}  
\usepackage{helvet}  
\usepackage{courier}  
\usepackage[hyphens]{url}  
\usepackage{graphicx} 
\usepackage{natbib}  
\usepackage{caption} 

\usepackage{subfiles}
\usepackage{booktabs}       
\usepackage{enumitem}
\usepackage{multirow}
\usepackage{amsmath}

\usepackage{algorithm}
\usepackage{algorithmic}

\title{RAG-Enhanced Collaborative LLM Agents for Drug Discovery}
\author{
    Namkyeong Lee\textsuperscript{\rm 1 2} \thanks{Work done while the author was an intern at Genentech.}, 
    Edward De Brouwer\textsuperscript{\rm 2}, 
    Ehsan Hajiramezanali\textsuperscript{\rm 2},\\
    Tommaso Biancalani\textsuperscript{\rm 2}, 
    Chanyoung Park\textsuperscript{\rm 1} \thanks{Corresponing Author}, 
    Gabriele Scalia\textsuperscript{\rm 2 $\dagger$}
}
\affiliations{
    \textsuperscript{\rm 1} KAIST, \{namkyeong96, cy.park\}@kaist.ac.kr \\
    \textsuperscript{\rm 2} Genentech, \{debroue1, hajiramm, biancalt, scaliag\}@gene.com
}

\usepackage{bibentry}

\usepackage{xspace}
\newcommand{\proposed}{\mbox{\textsf{CLADD}}\@\xspace}

\begin{document}

\maketitle

\begin{abstract}
Recent advances in large language models (LLMs) have shown great potential to accelerate drug discovery. However, the
specialized nature of biochemical data often necessitates costly domain-specific fine-tuning, posing major challenges. 
First, it hinders the application of more flexible general-purpose LLMs for cutting-edge drug discovery tasks. 
More importantly, it limits the rapid integration of the vast amounts of scientific data continuously generated through experiments and research. Compounding these challenges is the fact that real-world scientific questions are typically complex and open-ended, requiring reasoning beyond pattern matching or static knowledge retrieval.
To address these challenges, we propose \proposed, a retrieval-augmented generation (RAG)-empowered agentic system tailored to drug discovery tasks. Through the collaboration of multiple LLM agents, \proposed~dynamically retrieves information from biomedical knowledge bases, contextualizes query molecules, and integrates relevant evidence to generate responses --- all without the need for domain-specific fine-tuning. 
Crucially, we tackle key obstacles in applying RAG workflows to biochemical data, including data heterogeneity, ambiguity, and multi-source integration.
We demonstrate the flexibility and effectiveness of this framework across a variety of drug discovery tasks, showing that it outperforms general-purpose and domain-specific LLMs as well as traditional deep learning approaches.
Our code is publicly available at \textcolor{cyan}{\url{https://github.com/Genentech/CLADD}}.
\end{abstract}

\section{Introduction}
\label{sec: Introduction}
Large language models (LLM) have revolutionized the landscape of natural language processing, emerging as general-purpose foundation models with remarkable abilities across multiple  domains~\cite{achiam2023gpt,touvron2023llama}.
In particular, their application in biomolecular studies has recently gained significant interest, motivated by the potential to profoundly accelerate scientific innovation and drug discovery applications~\cite{zhang2024scientific,pei2024leveraging}. LLMs provide novel ways to understand and reason about molecular data, building on the wealth of available scientific literature. Additionally, their reasoning and zero-shot abilities help overcome the limitations of task-specific deep learning models, streamlining data needs and improving human-AI collaboration \cite{fang2023mol}. 

However, given the inherent complexity and specialized nature of the field, recent works emphasize the importance of domain-specific fine-tuning to boost tasks such as molecular captioning, property prediction, or binding affinity prediction~\cite{fang2023mol,chaves2024tx,yu2024llasmol,edwards2024molcap}. Consequently, rather than employing readily available general-purpose LLMs, most efforts in drug discovery have focused on fine-tuning LLMs using biochemical annotations or instruction-tuning datasets.

While promising, solely relying on these approaches poses significant challenges that can limit applications.
On one hand, the rapid emergence of new LLM architectures and techniques \cite{minaee2024large} complicates maintaining domain-specific models through expensive fine-tuning.
More importantly, drug discovery applications often require promptly incorporating new insights as they become available, for example, through new experiments or the scientific literature. 
In addition to being impractical, regular rounds of fine-tuning 
also introduce challenges such as catastrophic forgetting~\cite{luo2023empirical}, while not necessarily providing grounded answers~\cite{gekhman-etal-2024-fine}.
Above all, real-world drug discovery questions are inherently complex, open-ended, and context-dependent, spanning heterogeneous data types~\citep{ramos2025review}.
As a consequence, static LLMs---either general-purpose or fine‑tuned---may struggle to generalize to novel tasks or adapt to new evidence. 


From this perspective, retrieval-augmented generation (RAG) methods offer a promising direction that enables dynamic adaptation of the model's knowledge without the need for continuous, expensive fine-tuning \cite{gao2023retrieval,fan2024survey}.
However, applying this paradigm in the drug discovery domain presents important obstacles. First, retrieving relevant knowledge is difficult due to the limited domain expertise of general-purpose LLMs, combined with the vastness of the biochemical space \cite{bohacek1996art} that renders exact retrieval ineffective. 
Second, biochemical data is extremely heterogeneous, spanning diverse modalities such as molecules, proteins, diseases, and complex relationships between them~\cite{wang2023scientific}, which can also exist across multiple sources.
Finally, many real-world tasks are open-ended and require the LLM to extrapolate beyond the available external knowledge,
(which may also be ambiguous or partial \cite{vamathevan2019applications}) while remaining grounded in it. 

In this study, we tackle these challenges by introducing a \textbf{C}ollaborative framework of \textbf{L}LM \textbf{A}gents for \textbf{D}rug \textbf{D}iscovery (\proposed).
We assume a general setting where external knowledge is available as expert annotations associated with molecules or as knowledge graphs (KGs) that flexibly represent heterogeneous biochemical entities and their relationships. 
\proposed~is powered by general-purpose LLMs, while also integrating domain-specific LLMs, to improve molecular understanding. Notably, external knowledge can be updated dynamically without LLM fine-tuning. 

The multi-agent collaborative framework enables each agent to specialize in a specific data source and/or role, offering a modular solution that can improve overall information processing~\cite{chan2024chateval}. 
In particular, \proposed~includes a \emph{Planning Team} to determine relevant data sources, 
a \emph{Knowledge Graph Team} to retrieve  external heterogeneous information in the KG and summarize it, also through a novel anchoring approach to retrieve related information when the query molecule is not present in the knowledge base, 
and a \emph{Molecule Understanding Team}, which analyzes the query molecule based on its structure, along with summaries of external data and tools.
The flexibility of the framework enables \proposed~to address a wide range of tasks for drug discovery, including zero-shot and open-ended settings, while also improving interpretability through the transparent interaction of its agents.


Overall, we highlight the following contributions:
\begin{itemize}[leftmargin=.1in]
    \item We present \proposed, a multi-agent framework for RAG-based question-answering in drug discovery applications. The framework leverages generalist LLMs and dynamically integrates external biochemical data without fine-tuning, while addressing open-ended settings.
    \item We demonstrate the flexibility of the framework by tackling diverse applications, including drug-target prediction, property-specific molecular captioning, and biological activity prediction tasks. 
    \item We provide comprehensive experimental results showcasing the effectiveness of \proposed~compared to both general-purpose and domain-specific LLMs, as well as standard deep learning approaches. 
\end{itemize}

\begin{figure*}[t]
    \centering
    \includegraphics[width=0.80\linewidth]{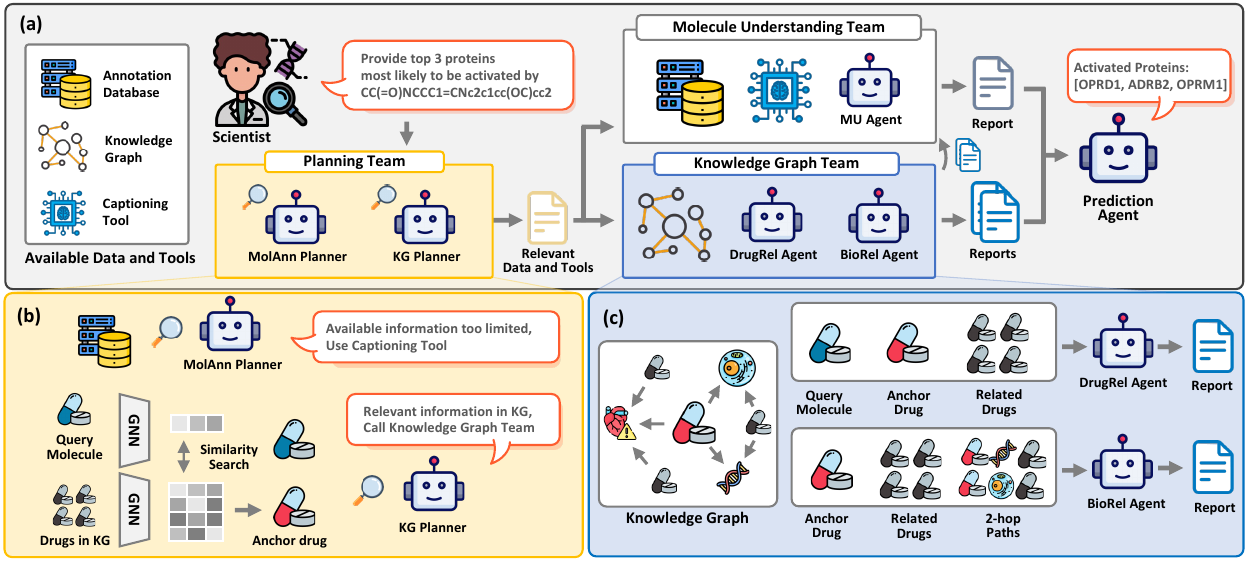}
    \vspace{-1ex}
    \caption{
    Overview of \proposed.
    }
    \label{fig:fig1}
    \vspace{-2ex}
\end{figure*}

\section{Related Work}
\label{sec: related_works}
\noindent \textbf{LLMs for Molecules.}
Leveraging the extensive body of literature and string-based molecular representations such as SMILES, language models (LMs) have been successfully applied to molecular sciences. 
Inspired by the masked language modeling approach used in BERT training \citep{devlin2018bert}, KV-PLM \citep{zeng2022deep} introduces a method to train LMs by reconstructing masked SMILES and textual data. 
Similarly, MolT5 \citep{edwards2022translation} adopts the ``replace corrupted spans" objective \citep{raffel2020exploring} for pre-training on both SMILES strings and textual data, followed by fine-tuning for downstream tasks such as molecule captioning and generation. 
Building on this foundation, \citet{pei2023biot5} and \citet{christofidellis2023unifying} extend MolT5 with additional pre-training tasks, including protein FASTA reconstruction and chemical reaction prediction.
Furthermore, GIMLET \cite{zhao2023gimlet}, Mol-Instructions \cite{fang2023mol}, and MolecularGPT \cite{liu2024moleculargpt} adopt instruction tuning \cite{zhang2023instruction} to improve generalization across a wide range of molecular tasks. 
While these approaches demonstrate enhanced versatility, they still rely on expensive fine-tuning processes to enable molecule-specific tasks or to incorporate new data.

\smallskip
\noindent \textbf{LLM Agents for Science.}
An LLM agent is a system that leverages LLMs to interact with users or other systems, perform tasks, and make decisions autonomously \cite{wang2024survey}. 
Recently, LLM agents have attracted significant interest in scientific applications and biomedical discovery~\cite{gao2024empowering}, with applications including literature search~\cite{lala2023paperqa}, experiment design~\cite{roohani2024biodiscoveryagent}, and hypothesis generation~\cite{wang-etal-2024-scimon}, among others. In particular, agents focusing on drug discovery applications have emerged. Systems like ChemCrow \cite{bran2023chemcrow}, CACTUS \cite{mcnaughton2024cactus}, and Coscientist \cite{boiko2023autonomous} focus on automating cheminformatics tasks and experiments, streamlining computational and experimental pipelines. Other works leverage agent-based orchestration of tools and data to accelerate specific aspects of scientific workflows, such as search~\cite{odonoghue2023bioplanner} or design~\cite{ghafarollahi2024protagents}. In contrast to existing works, we investigate an agent-based framework that can effectively incorporate external knowledge to improve open-ended and zero-shot molecular QA. This could be used either independently or as part of a larger system for automated drug discovery.


\smallskip
\noindent \textbf{Multi-Agent Collaborations for Drug Discovery.}
Only a limited number of studies have explored multi-agent frameworks in the context of drug discovery. 
DrugAgent~\cite{inoue2024drugagent} introduces a multi-agent framework integrating multiple external data sources, but is limited to predicting drug-target interaction.
Another study with the same name employs an agentic framework for automating machine learning programming for drug discovery tasks~\cite{liu2024drugagent}. 
In contrast, our work seeks to tackle a diverse array of drug discovery tasks, grounding the agent capabilities in external knowledge.
We provide further related works on knowledge graph-based RAG in Appendix \ref{app: Related Works}.


\section{Methodology}
\label{sec: Methodology}

\noindent \textbf{Problem Setup.}
Given a query molecule $g_q$ and a textual prompt describing a task of interest $\mathcal{I}$, we consider the general problem of generating a relevant response  $\mathcal{A}_{g_q}$. For instance, given $g_q=\text{`C1=CC(=C(C=C1CCN)O)O
'}$ and $\mathcal{I}=\text{`\emph{Predict liver toxicity}'}$, our model should be able to generate an answer stating that $\mathcal{A}_{g_q}=\text{`\emph{this molecule does not have liver toxicity concerns}'}$. 
Such a general QA setup can be flexibly adapted to multi-class classification, captioning, and set-based predictions. 
 
We assume access to two types of external databases: (1)~molecular annotation databases $\mathcal{C}$, which include textual annotation about molecules (for example, detailing their functions and properties), and (2)~knowledge graphs (KGs) connecting molecules to other biomedical entities.
In particular, a KG $\mathcal{G}$ is composed of a set of heterogeneous entities $\mathcal{E}$ (such as drugs, proteins, and diseases) and a set of relations $\mathcal{R}$ connecting them. 
In this paper, we only assume that molecule (or drug) entities are present in the KG, while any other types of entities can exist.
Additionally, we assume access to pre-trained molecular captioning models that can be used as external tools to complement the external databases. In general, any predictive model on molecules can be considered a captioning model \cite{edwards2022translation,pei2023biot5}, given that its output can be simply represented as text. 





\subsection{\proposed}

Here, we introduce \proposed, a multi-agent framework for general molecular question-answering that supports multiple drug discovery tasks. Each agent is implemented by an off-the-shelf LLM prompted to elicit a particular behavior.
Our framework is composed of three teams, each composed of several agents: 
the \textbf{Planning Team}, which identifies the most appropriate data sources and overall strategy given the task and the query molecule (Section~\ref{sec: Planning Team}); 
the \textbf{Knowledge Graph (KG) Team}, which retrieves relevant contextual information about the molecule from available KG databases (Section~\ref{sec:KnowledgeGraphTeam});
and the \textbf{Molecular Understanding (MU) Team}, which retrieves and integrates information from molecular annotation databases and external tools for molecule description (Section~\ref{sec: MolecualrUnderstandingTeam}). 
Finally, the \textbf{Prediction Agent}  integrates the findings from the MU and KG teams to generate the final answer.
In the following sub-sections, we describe each team in detail. The overall framework is depicted in Figure~\ref{fig:fig1}.

\subsubsection{Planning Team}
\label{sec: Planning Team}

The Planning Team assesses the relevance of external knowledge for a given query molecule. The team separately assesses the molecular annotations database and the knowledge graph through the MolAnn Planner and the KG Planner agents, respectively.

\smallskip
\noindent \underline{\textit{Molecule Annotation (MolAnn) Planner.}}~This agent first retrieves annotations for the query molecule, $c_q$, from the annotation database $\mathcal{C}$. While these annotations can provide valuable biochemical knowledge~\cite{yu2024llasmol}, they are often sparse, with many molecules entirely missing or lacking sufficient details due to the vastness of the chemical space~\cite{lee2024vision}.

To this end, the MolAnn Planner determines whether the retrieved annotations provide enough information for subsequent analyses.
Specifically, given a query molecule $g_q$, retrieved annotations $c_q$, and the task instruction $\mathcal{I}$, the agent is invoked as follows:
\begin{equation} 
\small
    o_{\text{MAP}} = \text{MolAnn Planner}(g_{q}, c_{q}, \mathcal{I}).
\end{equation}
$o_{\text{MAP}}$ indicates whether annotations should be complemented with additional information from tools.


\smallskip
\noindent \underline{\textit{Knowledge Graph (KG) Planner.}}~In parallel to analyzing the available description for the query molecule, we analyze the relevance of the contextual information present in the KG. 
While previous works on general QA tasks focus on identifying entities in the knowledge graph that exactly match those in the query~\cite{baek2023knowledge, jiang-etal-2023-structgpt}, the vast chemical search space and the limited coverage of existing knowledge bases limit the effectiveness of such approaches in the field of drug discovery.

To address this challenge, we propose leveraging the knowledge of drugs that are structurally similar to the query drug, building upon the well-established biochemical principle that structurally similar molecules often exhibit related biological activity \cite{martin2002structurally}. Specifically, we define the \emph{anchor drug} $g_{a}$ as the entity drug with the maximum cosine similarity between its embedding and that of the query molecule, among the set of all molecules in the KG ($g_{\mathcal{G}}$), $g_{a} = \underset{g \in g_{\mathcal{G}}}{\text{argmax}} \, \frac{\textit{emb}(g_q) \cdot \textit{emb}(g)} {\|\textit{emb}(g_q)\| \|\textit{emb}(g)\|}$, 
where $\textit{emb}$ is a representation produced by a graph neural network (GNN) pre-trained with 3D geometry~\cite{liu2021pre}, which outputs structure-aware molecular embeddings.

Then, the KG Planner agent decides whether to use the KG based on the structural similarity between the query molecule and the retrieved anchor drug.
To do so, we also provide the  Tanimoto similarity\footnote{We provide details on the Tanimoto similarity in Appendix \ref{app: Preliminaries}.}
to the KG Planner, as this domain-specific metric can be leveraged by  the LLM's reasoning about chemical structural similarity as follows:
\begin{equation} 
\small
    o_{\text{KGP}} = \text{KG Planner}(g_q, g_{a}, s_{q,a}, \mathcal{I}),
\end{equation}
where $s_{q,a}$ is the Tanimoto similarity between the query and anchor molecules.
$o_{\text{KGP}}$ is a Boolean indicating whether the KG should be used for the prediction. 

\subsubsection{Knowledge Graph Team}
\label{sec:KnowledgeGraphTeam}

This team aims to provide relevant contextual information about the query molecule by leveraging the KG, and it is only called if $o_{\text{KGP}} = \text{TRUE}$. 
It consists of the Drug Relation (DrugRel) Agent and the Biological Relation (BioRel) Agent, both of which generate reports on the query molecule based on different aspects of the KG. 
Specifically, the DrugRel Agent focuses on related drug entities within the KG, 
whereas the BioRel Agent focuses on summarizing and assessing contextual biological knowledge in the KG.

\smallskip
\noindent \underline{\textit{Related Drugs Retrieval.}}
The typical approach to leveraging a KG for QA tasks involves identifying multiple entities in the query and extracting the subgraph that encompasses those entities~\cite{wen2023mindmap}.
However, in molecular understanding for applications related to drug discovery tasks, the question often involves only a single entity, i.e., the query molecule $g_q$, making it challenging to identify information in the KG relevant to the task. 

Here, we introduce a novel approach for extracting relevant information for the query molecule $g_q$ by utilizing the retrieved anchor drug $g_a$, which exhibits high structural similarity to the query molecule.
In particular, while the drug entities in the KG $\mathcal{G}$ are mainly connected to other types of biological entities (e.g., proteins, diseases), we can infer relationships among drugs by considering the biological entities they share. 
For example, we can determine the relatedness of the drugs Trastuzumab and Lapatinib by observing their connectivity to the protein HER2 in the KG, as both drugs specifically target and inhibit HER2 to treat HER2-positive breast cancer~\cite{de2014lapatinib}.
Therefore, to identify relevant related drugs, we first compute the 2-hop paths connecting the anchor drug $g_{a}$ to other drugs $g_{\mathcal{G}}^{i}$ in the KG $\mathcal{G}$, i.e., $(g_{a}, r_{a \rightarrow e}, e, r_{i \rightarrow e}, g_{\mathcal{G}}^{i})$, where $r \in \mathcal{R}$, $e \in \mathcal{E}$, and $i$ denotes the index of the other drug.
Then, we select the top-$k$ \emph{related drugs}, denoted as $g_{r^{1}}, \ldots, g_{r^{k}}$, corresponding to the molecules that have the greatest number of 2-hop paths to the anchor drug.
Note that while the anchor drug $g_{a}$ is selected based on its structural similarity to the query molecule $g_{q}$, these reference drugs are \emph{semantically} related to $g_{a}$, reflecting the relationships captured within the KG.

\smallskip
\noindent \underline{\textit{Drug Relation (DrugRel) Agent.}}
The DrugRel Agent generates a report on the query molecule, contextualizing it in relation to relevant drugs present in the knowledge base for the specific task instruction.
Given a query molecule $g_q$, its anchor drug $g_{a}$, and the set of related drugs $g_{r^{1}}, \ldots, g_{r^{k}}$,
the DrugRel Agent generates a report as follows: 
\begin{equation} 
\small
    o_{\text{DRA}} = \text{DrugRel Agent}~(g_q, g_{a}, g_{r^{1}}, \ldots, g_{r^{k}}, \mathcal{T}, \mathcal{I}),
\end{equation}
where $\mathcal{T} = \{s_{q,a}, s_{q, r^{1}}, \ldots,  s_{q, r^{k}}\}$ is the set of Tanimoto similarities between the query molecule and the retrieved drugs.
The agent leverages its internal knowledge about related drugs while effectively assessing the relatedness of the information to the target molecule based on the Tanimoto similarity.

\smallskip
\noindent \underline{\textit{Biological Relation (BioRel) Agent.}}
The BioRel Agent summarizes how the anchor drug and the related drugs are biologically related, integrating additional biochemical entities present in the KG, such as targets, indications, side effects, etc.
Specifically, given an anchor drug $g_{a}$, a set of reference drugs $g_{r^{1}}, \ldots, g_{r^{k}}$, the collection of all 2-hop paths $\mathcal{P}$ linking the anchor drug to the reference drugs, and the instruction $\mathcal{I}$, the agent generates the report as follows:
\begin{equation} 
\small
    o_{\text{BRA}} = \text{BioRel Agent}(\mathcal{P}, \mathcal{I}, g_q, g_{a}, s_{q,a}).
    \vspace{-5pt}
\end{equation}
This enables us to obtain a task-relevant summary of the subgraph connected to the anchor drug. 

Importantly, while both the DrugRel Agent and BioRel Agent aim to reason about the query molecule in relation to other relevant entities in the KG for the specific task, they leverage distinct knowledge sources and perform different roles. 
Specifically, the BioRel Agent focuses on summarizing the network of relationships between drugs and other biological entities in the KG, contextualizing it with respect to the specific task at hand.
In contrast, the DrugRel Agent primarily draws on its internal knowledge, triggered by the names of the related drug entities in the KG, and incorporates structural similarity between them. 
In Section \ref{sec: Experiments}, we demonstrate how these agents complement each other, producing a synergistic effect when combined together.

\subsubsection{Molecular Understanding Team}
\label{sec: MolecualrUnderstandingTeam}

The Molecular Understanding (MU) Team compiles a report on the query molecule by leveraging external annotations and integrating them with structural information and reports from other agents.


\smallskip
\noindent \underline{\textit{Molecule Annotations.}}
Annotations from the external database are retrieved for the query molecule, denoted as $c_q$. If the Planning Team decided to use external annotation tools (i.e., $o_{\text{MAP}} = \text{TRUE}$), additional captions $ \Tilde{c}_q$ are generated with the external captioning tools as follows:
\begin{equation} 
\small
    \Tilde{c}_{q} = \text{Captioning Tools}(g_q),
\end{equation}
and concatenated to the annotations retrieved from the database: $c_{q} = c_{q} || \Tilde{c}_{q}$. 
External captioning tools allow the system to easily harness recent advances in LLM-driven molecular understanding~\cite{pei2023biot5,yu2024llasmol}, and can potentially include any tools, given that the output can be transformed into text.




\smallskip
\noindent \underline{\textit{Molecule Understanding (MU) Agent.}} The MU agent then analyzes the structure of the molecule, combining it with annotations and reports generated by the KG Team and generating a comprehensive report as follows: 
\begin{equation} 
\small
    o_{\text{MUA}} = \text{MU Agent}(g_{q}, c_{q}, o_{\text{DRA}}, o_{\text{BRA}}, \mathcal{I}).
\end{equation}

\subsubsection{Prediction Agent}
\label{sec:PredictionAgent}
Finally, the Prediction Agent performs the user-defined task by considering the reports from the various agents, including the MU and KG teams, 
as follows:
\begin{equation}
\small
    \mathcal{A}_{g_q} = \text{Task Agent}(g_q, o_{\text{MUA}}, o_{\text{DRA}}, o_{\text{BRA}}, \mathcal{I}).
\end{equation}
By integrating this evidence, the Prediction Agent can perform a comprehensive analysis of the query molecule. Importantly, the output of the Prediction Agent can be flexibly adjusted based on the specific task requirements. For instance, it can be a descriptive caption, a simple yes/no response for binary classification, or an open-ended answer. Such behavior leverages the zero-shot capabilities of LLMs~\cite{kojima2022large} and does not require additional fine-tuning.
Therefore, a key advantage of \proposed is its flexibility, which enhances scientist-AI interactions. 

\section{Experiments}
\label{sec: Experiments}

\smallskip
\noindent \textbf{Implementation Details.}
In all experiments, we utilize GPT-4o mini through the OpenAI API for each agent.
We use PrimeKG \cite{chandak2023building} as the KG, PubChem \cite{kim2021pubchem} as an annotation database, and MolT5 \cite{edwards2022translation} as an external captioning tool.
Additional implementation details and agent templates can be found in Appendix \ref{app: Implementation Details} and \ref{app: agent templates}, respectively.

\begin{table}[t]
    \small
    \centering
    \resizebox{0.95 \linewidth}{!}{
    \begin{tabular}{lcccccccc}
    \toprule
    & & \multicolumn{2}{c}{\textbf{(a) Overlap}} & & \multicolumn{2}{c}{\textbf{(b) No overlap}} \\
    \cmidrule{3-4} \cmidrule{6-7}
    & & \textbf{Activate} & \textbf{Inhibit} & & \textbf{Activate} & \textbf{Inhibit}\\ \midrule
    \textbf{GNNs (Fine-tune)} \\
    GraphMVP & & 1.76 & 1.03 & & 1.67 & 0.73\\
    MoleculeSTM & & 1.66 & 0.89 & & 1.48 & 0.65\\
    \midrule
    \multicolumn{3}{l}{\textbf{General LLMs (Zero-shot)}} \\
    GPT-4o mini & & 1.15 & 1.02 & & 1.13 & \underline{0.87} \\
    GPT-4o & & 0.62 & 0.79 & & 0.68 & 0.65 \\    
    \midrule
    \multicolumn{2}{l}{\textbf{Domain LMs (Zero-shot)}} & N/A & N/A & & N/A & N/A \\
    \midrule
    \multicolumn{3}{l}{\textbf{Domain LMs (Fine-tune)}}\\
    Galactica 125M & & 1.36 & 1.03 & & 0.86 & 0.69 \\
    Galactica 1.3B & & \underline{1.65} & \underline{1.09} & & \underline{1.37} & 0.80 \\
    Galactica 6.7B & & 1.52 & 0.97 & & 1.22 & 0.71 \\
    \midrule
    \proposed \textbf{(Zero-Shot)} & & \textbf{3.04} & \textbf{4.83} & & \textbf{2.67} & \textbf{3.24} \\
    \bottomrule
    \end{tabular}}
    \caption{Performance in drug-target prediction tasks (Precision @ 5). \textbf{Bold} and \underline{underline} indicate best and second-best language model-based methods.}
    \vspace{-3ex}
    \label{tab: protein target}
\end{table}

\subsection{Drug-Target Prediction Task}
\label{exp:Protein Target Prediction Task}

Accurately predicting a drug's protein target is essential for understanding its mechanism of action and optimizing its therapeutic efficacy while minimizing off-target effects \cite{santos2017comprehensive,batool2019structure}.
Here, we evaluate the models' ability to \emph{accurately identify which proteins a given molecule is most likely to activate or inhibit} in a set prediction setting. 

\smallskip
\noindent \textbf{Datasets.} We use molecular targets present in the Drug Repurposing Hub~\cite{corsello2017drug}, DrugBank~\cite{wishart2018drugbank}, and STITCH v5.0~\cite{szklarczyk2016stitch}, as preprocessed in \citet{zheng2023chempert}, including 13,688 molecules in total (details are presented in Appendix \ref{app: Datasets}). 

\smallskip
\noindent \textbf{Methods Compared.}
We evaluate two pre-trained GNNs, GraphMVP and MoleculeSTM, along with two general-purpose LLMs—GPT-4o mini and GPT-4o, and the domain-specific language model Galactica \cite{taylor2022galactica} (details are presented in Appendix \ref{app: Baselines}).

\smallskip
\noindent \textbf{Evaluation Protocol.}
We assess the performance of LLMs in a zero-shot setting. 
Specifically, for a given target molecule, we prompt the LLMs to generate the top 5 proteins that the molecule is most likely to activate or inhibit, and we calculate the precision with respect to ground truth data. 
As baseline GNNs cannot perform this task without training in a zero-shot setting, we fine-tune them in a few-shot setting using 10\% of the data. 
For domain-specific LMs, we also present fine-tuning results  on the specific task. 
To better assess generalization power, we separately report the performance on the test set for molecules present/not present in the external databases (``Overlap''/``No Overlap'').

\smallskip
\noindent \textbf{Experimental Results.}
Table~\ref{tab: protein target} summarizes the results. We observe the following:
\textbf{1)}~\proposed~outperforms all the baselines, with a higher likelihood of correctly identifying proteins activated/inhibited by the input molecule. 
\textbf{2)}~Importantly, the superiority of \proposed is confirmed for molecules not present in the caption database or knowledge graph (Table \ref{tab: protein target} (b)), showcasing \proposed's ability to leverage external knowledge to generalize to novel molecules.
\textbf{3)}~We observe that domain-specific fine-tuned models, such as Galactica, GIMLET, and MolecularGPT, \emph{could not perform this task in a zero-shot setting} when prompted to do so, likely because this task is not included in their fine-tuning instruction dataset.
By specifically fine-tuning Galactica on the task, we were able to answer the specific question, outperforming general-purpose LLMs in most experiments, but results were still inferior to \proposed.
This further highlights the flexibility of \proposed, which leverages the zero-shot abilities of general-purpose LLMs in its architecture. 

\begin{table}[t]
    \small
    \centering
    \resizebox{0.99 \linewidth}{!}{
    \begin{tabular}{lccccc}
    \toprule
    & & \textbf{BBBP} & \textbf{Sider} & \textbf{ClinTox} & \textbf{BACE}\\
    \midrule
    \textbf{GNNs} \\
    GraphMVP & & 69.59 \footnotesize{(1.29)} & 60.88 \footnotesize{(0.41)} & 87.57 \footnotesize{(3.26)} & 80.24 \footnotesize{(2.92)} \\
    MoleculeSTM & & 70.14 \footnotesize{(0.90)} & 58.69 \footnotesize{(0.89)} & 92.19 \footnotesize{(2.79)} & 79.24 \footnotesize{(3.40)} \\
    \midrule
    \textbf{Only SMILES} & & \underline{70.95} \footnotesize{(1.14)} & 60.80 \footnotesize{(1.18)} & 91.62 \footnotesize{(2.18)} & 74.21 \footnotesize{(1.32)} \\
    \midrule
    \textbf{General LLMs} \\
    GPT-4o mini & & 67.85 \footnotesize{(1.50)} & 58.18 \footnotesize{(1.55)} & 90.74 \footnotesize{(1.91)} & 74.22 \footnotesize{(1.95)} \\
    GPT-4o & & 66.43 \footnotesize{(1.47)} & 60.41 \footnotesize{(1.21)} & 88.13 \footnotesize{(1.74)} & 67.82 \footnotesize{(4.14)} \\
    \midrule
    \textbf{Domain LLMs} \\
    MolT5 & & 69.77 \footnotesize{(1.89)} & 57.20 \footnotesize{(0.98)} & 87.91 \footnotesize{(1.25)} & 74.28 \footnotesize{(4.00)} \\
    LlasMol & & 68.12 \footnotesize{(1.48)} & 61.50 \footnotesize{(1.66)} & 89.67 \footnotesize{(0.57)} & 75.42 \footnotesize{(2.98)} \\
    BioT5 & & 69.68 \footnotesize{(1.23)} & \underline{64.65} \footnotesize{(2.01)} & \underline{92.80} \footnotesize{(2.92)} & \underline{77.23} \footnotesize{(1.95)} \\        
    \midrule
    \proposed & & \textbf{72.28 \footnotesize{(1.04)}} & \textbf{66.42 \footnotesize{(1.31)}} & \textbf{93.80 \footnotesize{(2.30)}} & \textbf{77.74 \footnotesize{(3.15)}} \\
    \bottomrule
    \end{tabular}}
    \caption{Performance in molecular captioning tasks, mean AUROC with standard deviation (in parentheses). \textbf{Bold} and \underline{underline} indicate the best and second-best language model-based methods.}
    \vspace{-3ex}
    \label{tab: mol cap}
\end{table}

\subsection{Property-Specific Molecular Captioning Task}
\label{exp:Molecular Captioning Task}
Earlier studies on molecular captioning tasks have primarily focused on generating general descriptions of molecules without targeting specific areas of interest, raising concerns about their practical applicability in real-world drug discovery tasks.
Indeed, the usefulness of a molecular description is often task-dependent, and scientists may be interested in detailed explanations of specific characteristics of a molecule rather than a general description~\cite{guo2024moltailor,edwards2024molcap}.
Hence, in this paper, we introduce \emph{property-specific molecular captioning}, where the model is required to generate a description for a given molecule \emph{customized to a particular task of interest}.


\smallskip
\noindent \textbf{Datasets.}
We leverage four widely recognized molecular property prediction datasets from the MoleculeNet benchmark \cite{wu2018moleculenet}: \textbf{BBBP}, \textbf{Sider}, \textbf{ClinTox}, and \textbf{BACE} (further details in Appendix \ref{app: Datasets}).

\smallskip
\noindent \textbf{Methods Compared.}
We consider different baseline approaches. 
First, we compare recent molecular captioning methods designed to generate general descriptions of molecules, including MolT5~\cite{edwards2022translation}, LlasMol \cite{yu2024llasmol}, and BioT5~\cite{pei2023biot5}. Furthermore, we assess general-purpose LLMs, namely GPT-4o mini and GPT-4o.
Finally, for reference, we consider standard molecular property prediction baselines, including two GNNs pre-trained with different methodologies: GraphMVP \cite{liu2021pre} and MoleculeSTM \cite{liu2023multi}. 
We provide further details on the baseline models in Appendix~\ref{app: Baselines}.


\begin{table}[t]
    \small
    \centering
    \renewcommand*{\arraystretch}{0.8}
    \resizebox{0.99 \linewidth}{!}{
    \begin{tabular}{lcccccc}
    \toprule
    & & \multicolumn{4}{c}{\textbf{(a) Toxicity}} &\textbf{(b) \textbf{MLSMR}} \\
    \cmidrule{3-6}
    & & \textbf{hERG} & \textbf{DILI} & \textbf{Skin} & \textbf{Avg.} & \textbf{\textbf{Mtb}} \\
    \midrule
    \textbf{General LLMs} \\
    GPT-4o mini & & 28.42 & 33.47 & 41.84 & 34.58 & 33.33* \\
    GPT-4o & & 40.45 & 25.76 & \textbf{54.51} & 40.24 & 36.68  \\
    \midrule
    \textbf{Domain LLMs} \\
    Galactica 125M & & 40.78* & 33.56 & 42.43 & 38.92 & 33.33* \\
    Galactica 1.3B & & 48.57 & 34.37 & 42.43 & \underline{41.79} & 33.33*  \\
    Galactica 6.7B & & 23.75* & \underline{57.67} & 40.41* & 40.61 & 33.33* \\
    GIMLET & & 36.50 & 35.51 & 42.28 & 38.09 & \underline{39.81} \\
    LlasMol & & 23.75* & \textbf{61.20} & 31.92 & 38.95 & 33.33* \\
    \midrule
    \proposed & & \textbf{51.46} & 41.10 & \underline{50.43} & \textbf{47.66} & \textbf{50.92} \\
    \bottomrule
    \end{tabular}}
    \vspace{-1ex}
    \caption{Performance in biological activity prediction task including (a) toxicity and (b) antibacterial activity (Macro-F1). Avg. indicates the average performance over toxicity datasets. \textbf{Bold} and \underline{underline} indicate best and second-best methods.
    * indicates whether the model always outputs the same response, either ``Yes" or ``No".
    }
    \vspace{-3ex}
    \label{tab: toxicity}
\end{table}

\smallskip
\noindent \textbf{Evaluation Protocol.}
Although property-specific captions are practical, no ground truth property-specific captions exist for individual molecules, rendering traditional text generation evaluation methods inapplicable.
Thus, in line with recent works~\cite{xu2024llm,guo2024moltailor,edwards2024molcap}, we assess whether the generated captions can drive a classification model that categorizes molecules based on their properties.
Specifically, we pose this evaluation as a molecular property prediction problem, and fine-tune a SciBERT model~\cite{beltagy2019scibert} on the generated caption concatenated to the SMILES representation to predict the property of interest.
The ``Only SMILES" model utilizes only the SMILES string as input for the SciBERT classifier.
For baseline GNNs, each SMILES string is converted into a molecular graph. 
For all the experiments, we use a scaffold splitting strategy to simulate realistic distribution shifts, following previous work~\cite{liu2023multi}. 
This evaluation protocol is further illustrated in Appendix~\ref{app: Property-Specific Molecular Captioning Task}.

\smallskip
\noindent \textbf{Experimental Results.}
Table~\ref{tab: mol cap} summarizes the results.
\textbf{1)}~While domain-specific LLMs outperform general-purpose LLMs, their performance remains suboptimal, occasionally falling behind the ``Only SMILES" approach. This means that the generated captions occasionally reduce model performance compared to using only the SMILES representation of the molecule. This aligns with previous work that found that general descriptors may lack property-specific relevance~\cite{edwards2024molcap}.
\textbf{2)}~On the other hand, \proposed-generated captions consistently outperform all the baseline captioners and successfully improve over ``Only SMILES" across all datasets.
We attribute this improvement to the ability of \proposed to draw on external biochemical knowledge to ground its generation and its task-specificity. 
\textbf{3)}~Moreover, \proposed consistently outperforms pre-trained GNN baselines, except on the BACE dataset. 
Interestingly, this is also the only dataset for which the ``Only SMILES'' baseline falls short compared to GNN models, thus 
highlighting the critical role of 2D topological and 3D geometric information in this case.
This paves the way for future research on injecting essential aspects of molecules, such as geometric information, into LLM understanding.

\begin{figure}[t]
\centering
\includegraphics[width=0.9\linewidth]{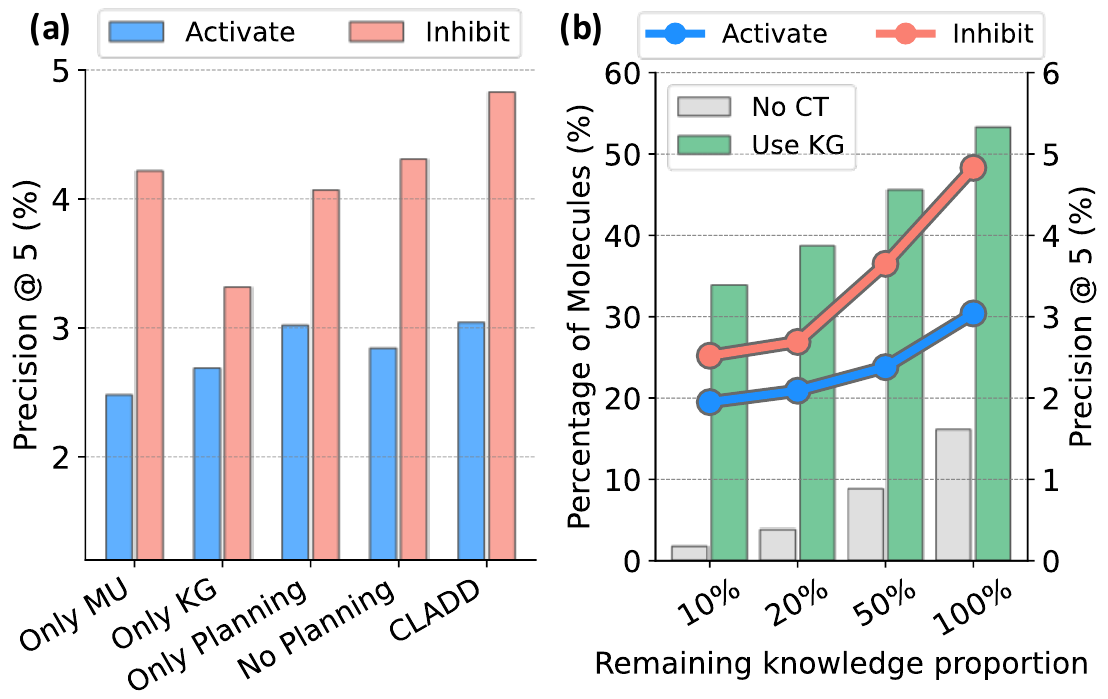}
    \caption{\textbf{Ablation studies.} \textbf{(a)} On model components. \textbf{(b)} On external knowledge.}
    \vspace{-2ex}
    \label{fig: model analysis}
\end{figure}

\begin{figure*}[t]
\centering
\includegraphics[width=0.9\linewidth]{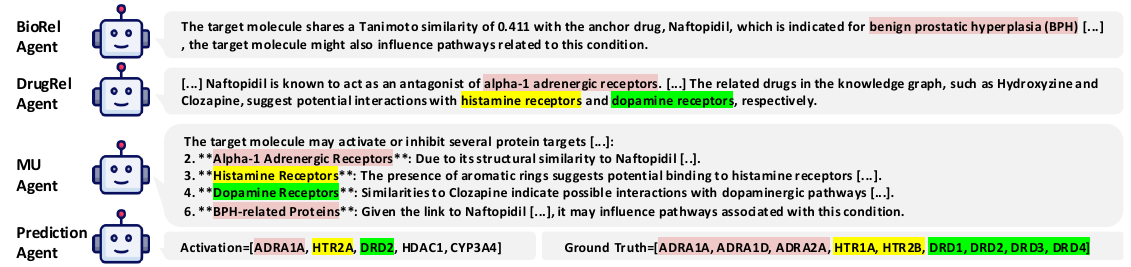}
    \caption{\textbf{Example of collaboration between agents in \proposed} (on the drug-target prediction task). Red represents adrenergic receptors, yellow represents histamine receptors, and green represents dopamine receptors (The full version in Appendix \ref{app: additional experiments}).
    }
    \vspace{-10pt}
    \label{fig: case studies}
\end{figure*}

\subsection{Biological Activity Prediction}
\label{exp:Drug Toxicity Prediction Task}

Accurately predicting molecular bioactivity is a cornerstone of drug discovery, which is often hindered by the existence of countless biological contexts and sparse experimental data. 
We therefore explore the \emph{zero-shot characterization of biological activity for unseen compounds}. To this goal, we focus on  \emph{drug toxicity}~\cite{basile2019artificial} and \emph{antibacterial activity}~\cite{melo2021accelerating} prediction. 

\smallskip
\noindent \textbf{Datasets.}
For drug toxicity prediction, we use three benchmark datasets: \textbf{hERG}~\cite{wang2016admet}, \textbf{DILI}~\cite{xu2015deep}, and \textbf{Skin}~\cite{alves2015predicting}. For antibacterial activity prediction, we use a dataset curated from \citet{eke2025genetic}, hereafter referred to as \textbf{MLSMR\_Mtb}. In addition to its relevance, we selected MLSMR\_Mtb for its recency, as it was \emph{published after GPT-4o training and in parallel to the preparation of this study}, therefore avoiding the risk of pre-training data leakage.
Dataset details are presented in Appendix~\ref{app: Datasets}.

\smallskip
\noindent \textbf{Methods Compared.}
We compare five domain-specific LLMs—Galactica 125M, Galactica 1.3B, Galactica 6.7B \cite{taylor2022galactica}, LlasMol \cite{yu2024llasmol}, and GIMLET \cite{zhao2023gimlet}, alongside two general-purpose LLMs, GPT-4o and GPT-4o mini (details in Appendix \ref{app: Baselines}).

\smallskip
\noindent \textbf{Evaluation Protocol.}
Evaluation follows a zero-shot QA setting. The input includes a SMILES representation of the molecule and the task description.  
Using the text-formatted output generated by each model, we compute the Macro-F1 score \cite{opitz2019macro} as the evaluation metric.

\smallskip
\noindent \textbf{Experimental Results.}
Table~\ref{tab: toxicity} summarizes the results.
\textbf{1)}~Both on toxicity datasets (average score) and the recently published antibacterial activity dataset,  \proposed~outperforms all the baselines. This includes GPT-4o mini, which is used as building block of \proposed. This highlights its ability to perform zero-shot predictions without domain-specific fine-tuning by effectively incorporating external knowledge into general-purpose LLMs at inference time.
\textbf{2)}~
Notably, for three datasets (hERG, Skin and MLSMR\_Mtb), several baseline models often output the same response, either ``Yes" or ``No", indicating their inability to perform the given task. In contrast, \proposed~did not suffer from this limitation. \textbf{3)}~Most baselines performed especially poorly on the recently released dataset (MLSMR\_Mtb).
However, \proposed~shows a significant improvement over the baselines, demonstrating its reasoning ability on unseen tasks that are even published after the language model.


\subsection{Ablation studies}


\smallskip
\noindent \textbf{Model Components Ablations.}
In Figure~\ref{fig: model analysis} (a), we report the results of ablations on the components of~\proposed. We observe:
\textbf{1)}~\emph{The knowledge graph and the molecular annotations are important and complementary data sources}, as shown by the lower performance when only Molecular Understanding or Knowledge Graph team is available (``Only MU'', ``Only KG''). 
\textbf{2)}~\emph{Dynamically selecting the relevant data sources with Planning Team improves performance}, leveraging their complementarity, as suggested by the lower performance of the ``No Planning''.
\textbf{3)} 
\emph{The distributed architecture of the multi-agent system is a more effective way of processing the retrieved information}, as highlighted by the lower performance of ``Only Planning'' where all the relevant data sources are directly included in the prompt of a single Prediction Agent, bypassing intermediate reports. 
Additional ablation studies are presented in Appendix~\ref{app: additional ablation studies}. Furthermore, \emph{we confirmed results across different LLMs, including open-source models}, showcasing the LLM-agnostic nature of \proposed in Appendix~\ref{app: more llms}.

\smallskip
\noindent \textbf{External Knowledge Ablations.}
To further assess the impact of external knowledge on model performance, we evaluate the model after progressively pruning the available databases and present our results in Figure \ref{fig: model analysis} (b). We observe the following:
\textbf{1)} \emph{Model performance depends on external knowledge size}, validating the key role of the external knowledge to the framework.  \textbf{2)} Interestingly, \emph{we do not observe any performance plateau}, indicating that further expanding the external knowledge could provide additional performance improvements.
\textbf{3)} From the bar plots, i.e., ``No CT (No Captioning Tool)" and ``Use KG (Call Knowledge Graph Team)", we observe that as the amount of external knowledge grows, the planning team increasingly depends on it. 
This indicates that \proposed~\emph{actively leverages external knowledge more effectively during the decision-making process when such knowledge is more abundant}.
A more detailed analysis of how external knowledge is utilized and its impact on model performance is provided in Appendix \ref{app: additional external knowledge analysis}.

\vspace{-1ex}
\subsection{Case Studies}

Figure \ref{fig: case studies} showcases how the agents in \proposed collaborate to identify ``the top-5 protein targets a query molecule is most likely to activate''. 
First, the BioRel Agent extracts from the knowledge graph that the anchor drug, Naftopidil, is indicated for benign prostatic hyperplasia (BPH), implying the activation of related pathways. 
The DrugRel Agent complements these findings by 
\textbf{1)}~linking BPH to alpha-1 adrenergic receptors using its internal knowledge (which is confirmed in the literature~\cite{klotsman2004case}), and 
\textbf{2)}~analyzing related drugs in the knowledge graph (\emph{e.g.,} Hydroxyzine,  Clopamine), to infer interactions with histamine and dopamine receptors. 
Finally, the MU agent integrates these findings with the analysis of the molecular structure to provide a summarized report of the activated protein targets. 
This example highlights the agents' complementary strengths, which lead to interpretable and reliable predictions. 
Additional case studies
are provided in Appendix~\ref{app: additional case studies}.


\section{Conclusion}
In this work, we introduced \proposed, a RAG-enhanced multi-agent framework for zero-shot molecular question-answering that can support various drug discovery tasks. We showcased its flexibility and effectiveness across multiple real-world tasks, outperforming both general-purpose and domain-specific fine-tuned LLMs. Our analyses highlighted the complementarity of external knowledge sources, internal LLM reasoning, and multi-agent orchestration. \proposed's chain of messages also provides insight into its decision-making process, fostering more interpretable interactions.

\section*{Acknowledgement}
N.L. and C.P. were supported by the Institute of Information \& Communications Technology Planning \& Evaluation (IITP) grant funded by the Korea government (MSIT) (RS-2025-02304967, AI Star Fellowship (KAIST)). Additionally, N.L. and C.P. received funding from the National Research Foundation of Korea (NRF) through two separate grants: RS-2024-00335098 (funded by the Korea government (MSIT)) and RS-2022-NR068758 (funded by the Ministry of Science and ICT).

E.D.B., E.H., T.B., G.S. are employees of Genentech and shareholders of Roche.

\bibliography{aaai2026}

\clearpage
\appendix

\onecolumn

\section{Additional Related Works}
\label{app: Related Works}
\noindent \textbf{LLMs with Knowledge Graphs.}
While large language models (LLMs) have been successfully adapted to numerous domains, they have faced criticism for their lack of factual accuracy. Specifically, LLMs often struggle to recall reliable facts and are prone to hallucinations~\cite{ji2023survey}, which can be a bottleneck for scientific applications, and are still persistent after fine-tuning~\cite{gekhman-etal-2024-fine}.
A promising approach to mitigate these issues is the integration of external knowledge sources, such as knowledge graphs (KGs), into LLMs during the generation process. 
For instance, \citet{baek2023knowledge} proposes a method where relevant triplets are retrieved from KGs based on the input query. These triplets are then verbalized and provided as additional input to the LLM, enhancing its factual grounding and accuracy.
KG-Rank~\cite{yang2024kg} focuses on medical question-answering, leveraging a medical knowledge graph to match terms in the question and expand them. DALK~\cite{li2024dalk} leverages an LLM to construct an Alzheimer's disease-specific KG, which is then used to enhance the accuracy and relevance of LLM-generated responses.
Although these methods retrieve entities from KGs that are related to those in the query, the virtually infinite number of potential molecules of interest in drug discovery, combined with the limited domain expertise of general-purpose LLMs, makes it challenging to directly apply existing techniques to molecular question-answering.

\section{Preliminaries}
\label{app: Preliminaries}




\noindent \textbf{Tanimoto Similarity.}
The Tanimoto similarity is a widely accepted criterion for calculating the similarity between two molecules based on their molecular fingerprint \cite{bajusz2015tanimoto}, which are the binary sequences that denote the presence or absence of specific substructures \cite{rogers2010extended}.
Given two molecules $g_i$ and $g_j$ with fingerprints $\mathtt{fp}_i$ and $\mathtt{fp}_j$, the Tanimoto similarity $s_{i,j}$ is computed as follows:
\begin{equation}
\small
s_{i,j} = \frac{|\mathtt{fp}_i \cap \mathtt{fp}_j|}{|\mathtt{fp}_i| + |\mathtt{fp}_j| -  |\mathtt{fp}_i \cap \mathtt{fp}_j|}.
\label{eq: Tanimoto}
\end{equation}
Intuitively, the Tanimoto similarity is the intersection-over-union of the sets of molecular substructures of both molecules.


\section{Datasets}
\label{app: Datasets}
In this section, we provide further details on the datasets we used in Section \ref{sec: Experiments}.
We provide a summary of data statistics in Table \ref{app tab: data stats}.

\begin{table}[h!]
    \centering
    \caption{Data statistics.}
    \resizebox{0.8\linewidth}{!}{
    \begin{tabular}{lccccccccccc}
    \toprule
    & \multirow{2}{*}{\textbf{hERG}} & \multirow{2}{*}{\textbf{DILI}} & \multirow{2}{*}{\textbf{Skin}} & \multirow{2}{*}{\textbf{MLSMR\_Mtb}} & \multirow{2}{*}{\textbf{BBBP}} & \multirow{2}{*}{\textbf{Sider}} & \multirow{2}{*}{\textbf{ClinTox}} & \multirow{2}{*}{\textbf{BACE}} & & \multicolumn{2}{c}{\textbf{ChemPert}}\\
    \cmidrule{11-12}
    &  &  &  &  &  &  &  &  &  & Overlap & No Overlap \\
    \midrule
    \# Molecules & 648 & 475 & 404 & 200 & 2039 & 1427 & 1477 & 1513 &  & 7917 & 5771\\
    \midrule
    \# Tasks & 1 & 1 & 1 & 1 & 1 & 27 & 2 & 1 &  & 2 & 2\\
    \bottomrule
    \end{tabular}}
    \label{app tab: data stats}
\end{table}

\subsection{Drug Biological Activity Prediction Task}
For the drug biological activity prediction task, we use four datasets: \textbf{hERG}, \textbf{DILI}, \textbf{Skin}, and \textbf{MLSMR\_Mtb}.
\begin{itemize}[leftmargin=.1in]
\item The Human ether-a-go-go related gene (\textbf{hERG}) \cite{wang2016admet} plays a critical role in regulating the heart's rhythm. Thus, accurately predicting hERG liability is essential in drug discovery. In this task, we assess the model's ability to predict whether a drug blocks hERG.
\item Drug-induced liver injury (\textbf{DILI}) \cite{xu2015deep} is a severe liver condition caused by medications. In this task, we evaluate the model's capability to predict whether a drug is likely to cause liver injury.
\item Repeated exposure to a chemical agent can trigger an immune response in inherently susceptible individuals, resulting in \textbf{Skin} \cite{alves2015predicting} sensitization. In this task, we evaluate the model's ability to predict whether the drug induces a skin reaction.
\item The Molecular Libraries Small Molecule Repository - \emph{Mycobacterium tuberculosis} dataset (\textbf{MLSMR\_Mtb}) has been released as part of \citet{eke2025genetic}. Antimycobacterial activity against \emph{M.~tuberculosis} was measured in a dose-response assay and quantified as AUC. Following the original study, we used an AUC cutoff of 25 for classification. Out of the 935 molecules tested, we randomly selected 200 compounds with a balanced positive/negative ratio. For this task, we evaluate the model's ability to predict antimycobacterial activity. 
In addition to its relevance, we selected this dataset for its recency, as \textbf{it was published after GPT-4o and in parallel to the preparation of this study}, ensuring no overlap with pre-training data and thus allowing benchmarking against leakage risks. To the best of our knowledge, our work is the first study leveraging this dataset. 

\end{itemize}

\subsection{Property-Specific Molecular Captioning Task}
\label{app: Property-Specific Molecular Captioning Task}
For the property-specific molecular captioning task, we use four datasets in MoleculeNet \cite{wu2018moleculenet}: \textbf{BBBP}, \textbf{Sider}, \textbf{Clintox}, \textbf{BACE}.
\begin{itemize}[leftmargin=.1in]
\item The blood-brain barrier penetration \textbf{(BBBP)} dataset consists of compounds categorized by their ability to penetrate the barrier, addressing a significant challenge in developing drugs targeting the central nervous system.
\item The side effect resource \textbf{(Sider)} dataset organizes the side effects of approved drugs into 27 distinct organ system categories.
\item The \textbf{Clintox} dataset includes two classification tasks: 1) predicting toxicity observed during clinical trials, and 2) determining FDA approval status.
\item The \textbf{BACE} dataset provides qualitative binding results for a set of inhibitors aimed at human $\beta$-secretase 1.
\end{itemize}

\begin{figure}[h!]
\centering
\includegraphics[width=0.5\linewidth]{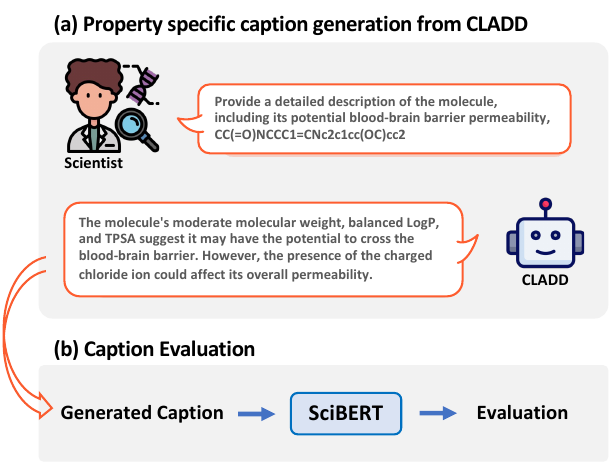}
    \caption{(a) After CLADD (or baseline models) generates a property-specific caption, (b) SciBERT is used for evaluation. In other words, \textbf{fine-tuning SciBERT is not part of CLADD}; it is only used for evaluation purposes.
    }
    \label{fig: evaluation protocol}
\end{figure}

\smallskip
\noindent\textbf{Evaluation Protocol.} 
While previous works on molecular captioning generate general molecule descriptions and evaluate them with standard NLP metrics like BLEU. 
However, because a molecule can be described in multiple ways (some more relevant to certain tasks~\cite{guo2024moltailor,edwards2024molcap}), we focus on property-specific captioning. 
Here, the main challenge is the lack of ground-truth captions for each property. 
Therefore, similar to previous work \cite{edwards2024molcap}, we use an evaluation protocol that checks how well the generated captions aid in property prediction by fine-tuning a language model (SciBERT) on them. Specifically, for a generated \texttt{caption} and the \texttt{SMILES} representation of the target molecule, we concatenate them using a \texttt{[CLS]} token, forming \texttt{SMILES[CLS]caption}, and fine-tune a SciBERT \cite{beltagy2019scibert} model for property prediction. Importantly, \textbf{fine-tuning SciBERT is only part of the evaluation protocol, as CLADD itself does not involve any fine-tuning}. This process is illustrated in  Figure~\ref{fig: evaluation protocol}.

\subsection{Drug-Target Prediction Task}
We rely on annotated molecular targets present in the Drug Repurposing Hub~\cite{corsello2017drug}, DrugBank~\cite{wishart2018drugbank}, and STITCH v5.0~\cite{szklarczyk2016stitch}, as combined and preprocessed in \citealp{zheng2023chempert}. 
As we explained in Section \ref{sec: Experiments}, 
we separately report the performance on the test set for molecules based on their information availability in the external databases (``Overlap''/``No Overlap'').
More specifically, for ``No Overlap" cases, we exclude the molecules in the following criteria:
\begin{itemize}[leftmargin=.1in]
\item We exclude the molecules if they exist in the knowledge graph.
\item However, we noticed that many molecules have uninformative annotations, as also discussed in Section \ref{app: Implementation Details}. Consequently, we decided to exclude molecules from the test set only if they have sufficient annotations relevant to the task, as determined by GPT-4o mini.
\end{itemize}
After this process, 5771 molecules remained in the test set for the ``No Overlap" scenario.


\section{Baselines Setup}
\label{app: Baselines}
This section provides further details on the baselines we used in Section \ref{sec: Experiments}.
For all baseline models, we utilize the pre-trained checkpoints provided by the authors of the original papers.
\begin{table}[h!]
    \centering
    \caption{Links to baseline model checkpoints.}
    \resizebox{0.5\linewidth}{!}{
    \begin{tabular}{l|l}
    \toprule
    Model & URL \\
    \midrule
    Galactica 125M & \url{https://huggingface.co/facebook/galactica-125m}\\
    Galactica 1.3B& \url{https://huggingface.co/facebook/galactica-1.3b}\\
    Galactica 6.7B& \url{https://huggingface.co/facebook/galactica-6.7b}\\
    GIMLET & \url{https://huggingface.co/haitengzhao/gimlet}\\
    LlasMol & \url{https://huggingface.co/osunlp/LlaSMol-Mistral-7B}\\
    MolecularGPT & \url{https://huggingface.co/YuyanLiu/MolecularGPT}\\
    \bottomrule
    \end{tabular}}
    \label{app tab: baselines}
\end{table}

\begin{itemize}[leftmargin=.1in]
\item \textbf{Galactica} \cite{taylor2022galactica} is a large language model designed to store, integrate, and reason over scientific knowledge. The authors demonstrate Galactica's capabilities in simple molecule understanding tasks, such as predicting IUPAC names and performing binary classification for molecular property prediction.
We also fine-tune Galactica for the Drug-Target Prediction task described in Section~\ref{sec: Experiments}, using molecules and associated activated/inhibited proteins. 
For fine-tuning, we searched for the optimal hyperparameters (learning rate of $\{1e-3, 1e-4, 1e-5, 1e-6\}$ and epoch number of $\{50, 100, 150, 200\}$), reporting the best performance achieved.

\item \textbf{GIMLET} \cite{zhao2023gimlet} introduces a unified approach to leveraging language models for both graph and text data. The authors aim to enhance the generalization ability of language models for molecular property prediction through instruction tuning.
\item \textbf{LlaSMol} \cite{yu2024llasmol} presents a large-scale, comprehensive, and high-quality dataset designed for instruction tuning of large language models. This dataset includes tasks such as name conversion, molecule description, property prediction, and chemical reaction prediction, and it is used to fine-tune different open-source LLMs.
\end{itemize}

\section{Implementation Details}
\label{app: Implementation Details}
In this section, we provide further details on the implementation of \proposed.

\smallskip
\noindent\textbf{Software Configuration.}
Our model is implemented using Python 3.11, PyTorch 2.5.1, Torch-Geometric 2.6.1, RDKit 2023.9.6, and LangGraph 0.2.59.

\smallskip
\noindent \textbf{Computational Resources.}
For LLMs, we utilize the OpenAI API, thereby leveraging OpenAI’s computational resources. All other computations, such as GNN retrievers, are performed on a 24GB NVIDIA GeForce RTX 3090 GPU.

\smallskip
\noindent\textbf{External Databases.}
In all experiments, we employ the PubChem database \cite{kim2021pubchem} as the annotation database $\mathcal{C}$ and PrimeKG \cite{chandak2023building} as the biological knowledge graph $\mathcal{G}$.

\begin{figure*}[t]
    \centering
    \begin{minipage}{0.5\linewidth}
        \centering
        \includegraphics[width=0.95\linewidth]{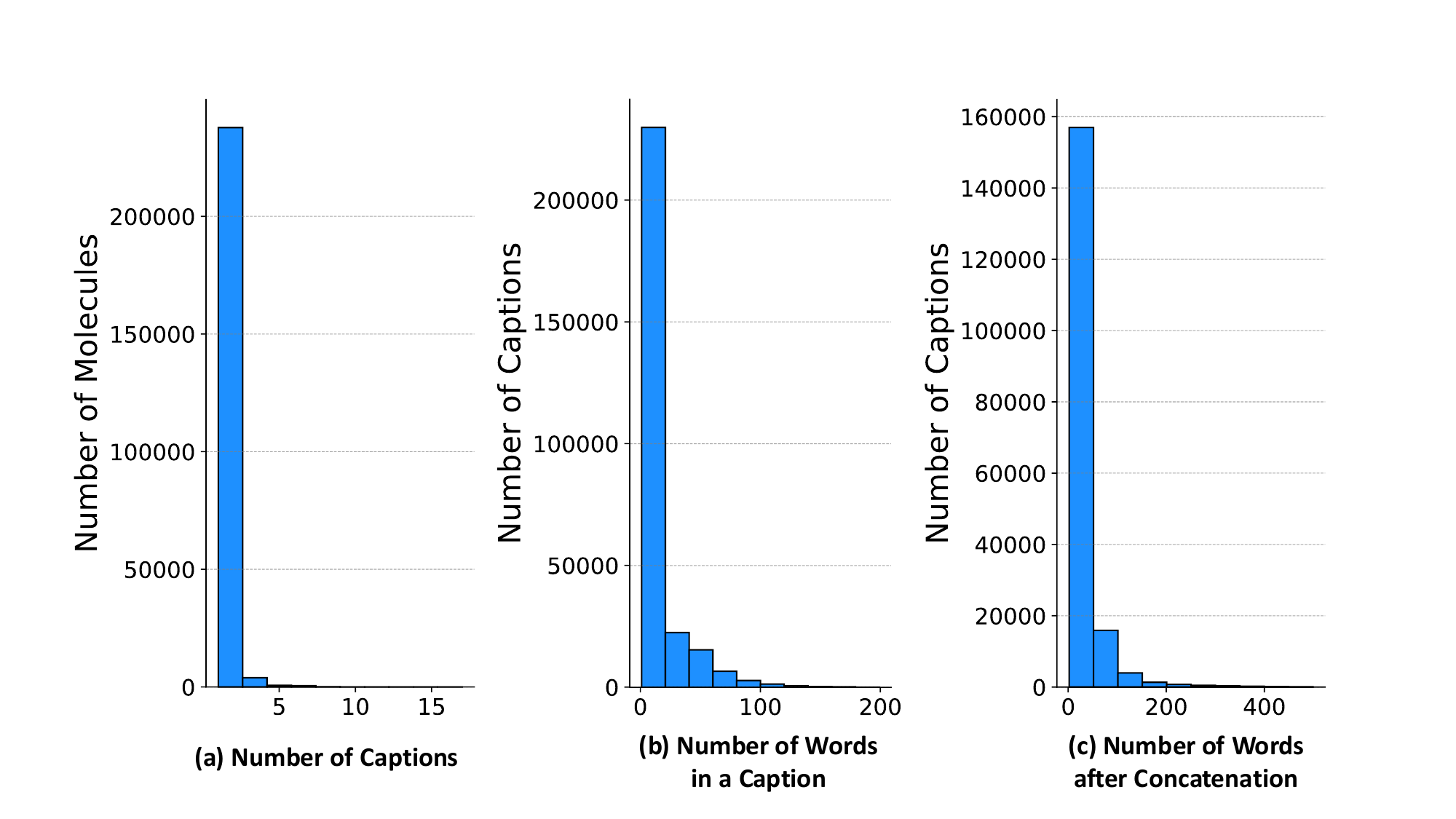}
        \caption{Data analysis on PubChem database.}
        \label{app fig: data analysis pubchem}
    \end{minipage}
    \begin{minipage}{0.49\linewidth}
        \centering
        \includegraphics[width=0.95\linewidth]{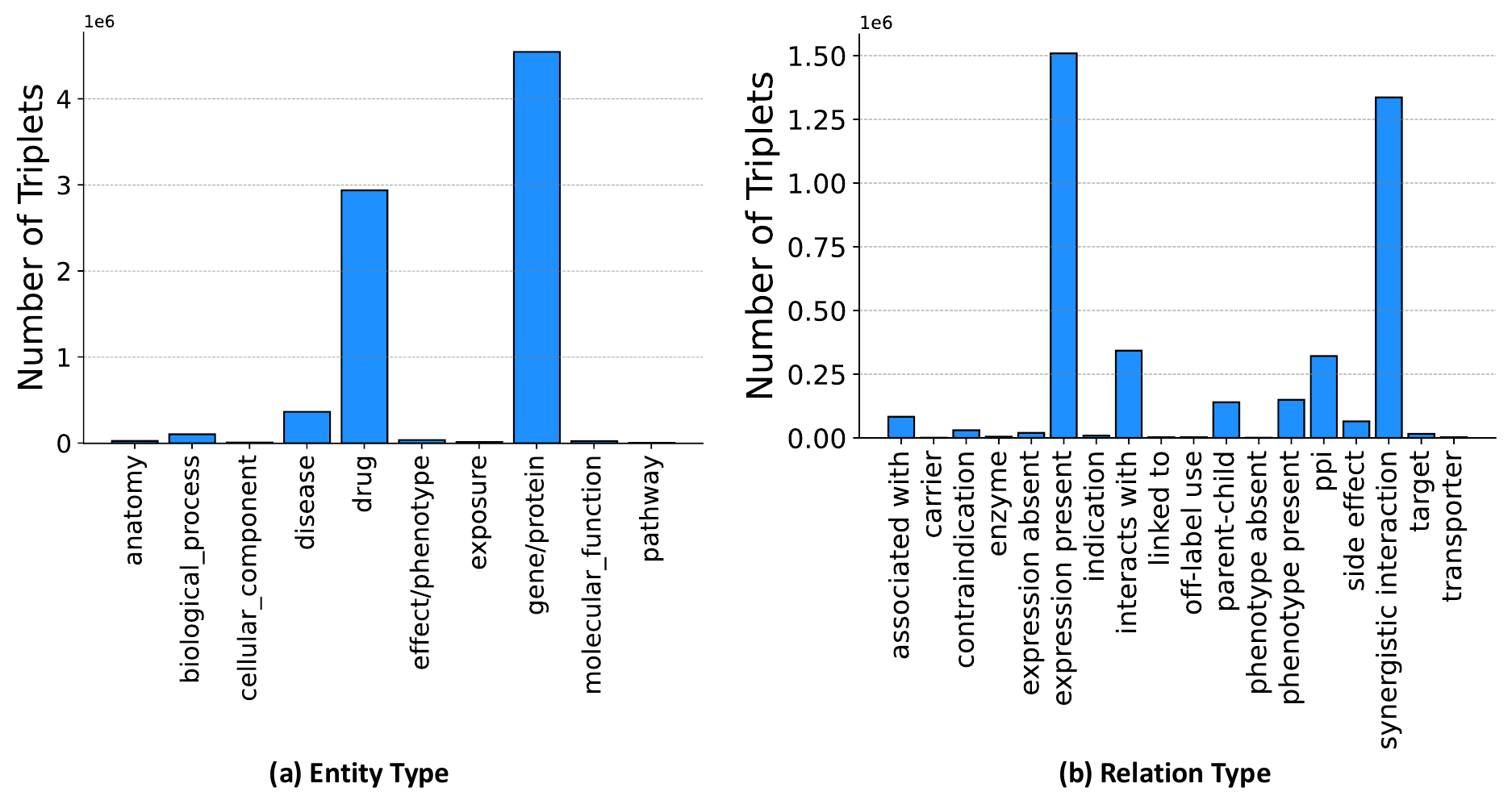}
        \caption{Data analysis on PrimeKG knowledge graph.}
        \label{app fig: data analysis primekg}
    \end{minipage}
\end{figure*}

The \textbf{PubChem} database is one of the most extensive public molecular databases available.
Pubchem database consists of multiple data sources, including DrugBank, CTD, PharmGKB, and more (\url{https://pubchem.ncbi.nlm.nih.gov/sources/)}.
The PubChem database used in this study includes 299K unique molecules and 336K textual descriptions associated with them (that is, a single molecule can have multiple captions sourced from different datasets associated with it). 
On average, each molecule has 1.115 descriptions, ranging from a minimum of one to a maximum of 17, as shown in Figure \ref{app fig: data analysis pubchem} (a).
In this study, if a molecule had multiple captions, they were concatenated to form a single caption.
On the other hand, as shown in Figure \ref{app fig: data analysis pubchem} (b), most captions consist of fewer than 20 words, underscoring the limited informativeness of human-generated captions. Even after concatenating multiple captions for each molecule, the majority still contain fewer than 50 words.

\textbf{PrimeKG} is a widely used knowledge graph for biochemical research.
The knowledge graph contains 4,037,851 triplets and encompasses 10 entity types, including \{\texttt{anatomy, biological processes, cellular components, diseases, drugs, effects/phenotypes, exposures, genes/proteins, molecular functions, and pathways}\}. 
Additionally, it includes 18 relationship types: \{\texttt{associated with, carrier, contraindication, enzyme, expression absent, expression present, indication, interacts with, linked to, off-label use, parent-child, phenotype absent, phenotype present, ppi, side effect, synergistic interaction, target, and transporter}\}.
The number of triplets associated with each entity and relation type is shown in Figure \ref{app fig: data analysis primekg} (a) and (b), respectively.

\subsection{KG Planner}
As explained in section \ref{sec: Planning Team}, we utilize a pre-trained GNN (with 3D information) to retrieve molecules highly related to the query molecule.
In particular, the model has a GIN architecture \cite{xu2018powerful}, which is pre-trained with the GraphMVP \cite{liu2021pre} approach.
The checkpoint of the model is available at~\url{https://huggingface.co/chao1224/MoleculeSTM/tree/main/pretrained_GraphMVP}.


\section{Additional Experimental Results}
\label{app: additional experiments}
In this section, we provide additional experimental results that can supplement our experimental results in Section \ref{sec: Experiments}.

\subsection{Additional Ablation Studies}
\label{app: additional ablation studies}
\begin{figure}[h]
    \centering
    \begin{minipage}{0.54\linewidth}
    \centering
    \captionof{table}{Additional ablation studies in drug-target prediction task (Precision @ 5). \textbf{Bold} and \underline{underline} indicate best and second-best methods.}
    \resizebox{0.95\linewidth}{!}{
    \begin{tabular}{lcccccccc}
    \toprule
    & & \multicolumn{2}{c}{\textbf{(a) Overlap}} & & \multicolumn{2}{c}{\textbf{(b) No overlap}} \\
    \cmidrule{3-4} \cmidrule{6-7}
    & & \textbf{Activate} & \textbf{Inhibit} & & \textbf{Activate} & \textbf{Inhibit}\\ \midrule
    \textbf{No MolAnn Planner} \\
    - Only Expert Annotation  & & 2.99 & 4.80 & & 2.63 & \underline{3.20} \\ 
    - Only Generated Caption  & & 2.72 & 3.96 & & 2.61 & 2.80 \\ 
    \textbf{No KG Planner} & & 2.84 & 4.49 & & 2.64 & 2.97 \\ 
    \textbf{No DrugRel Agent} & & 2.90 & \underline{4.79} & & 2.48 & 2.99 \\ 
    \textbf{No BioRel Agent} & & 2.96 & 4.50 & & 2.63 & 3.00 \\ 
    \textbf{No MU Agent} & & \textbf{3.04} & 4.17 & & \underline{2.66} & 2.59 \\ 
    \midrule
    \proposed & & \textbf{3.04} & \textbf{4.83} & & \textbf{2.67} & \textbf{3.24} \\
    \bottomrule
    \end{tabular}}
    \label{app tab: ablations}
    \end{minipage}
    \begin{minipage}{0.45\linewidth}
        \centering
        \vspace{4ex}
        \includegraphics[width=0.95\linewidth]{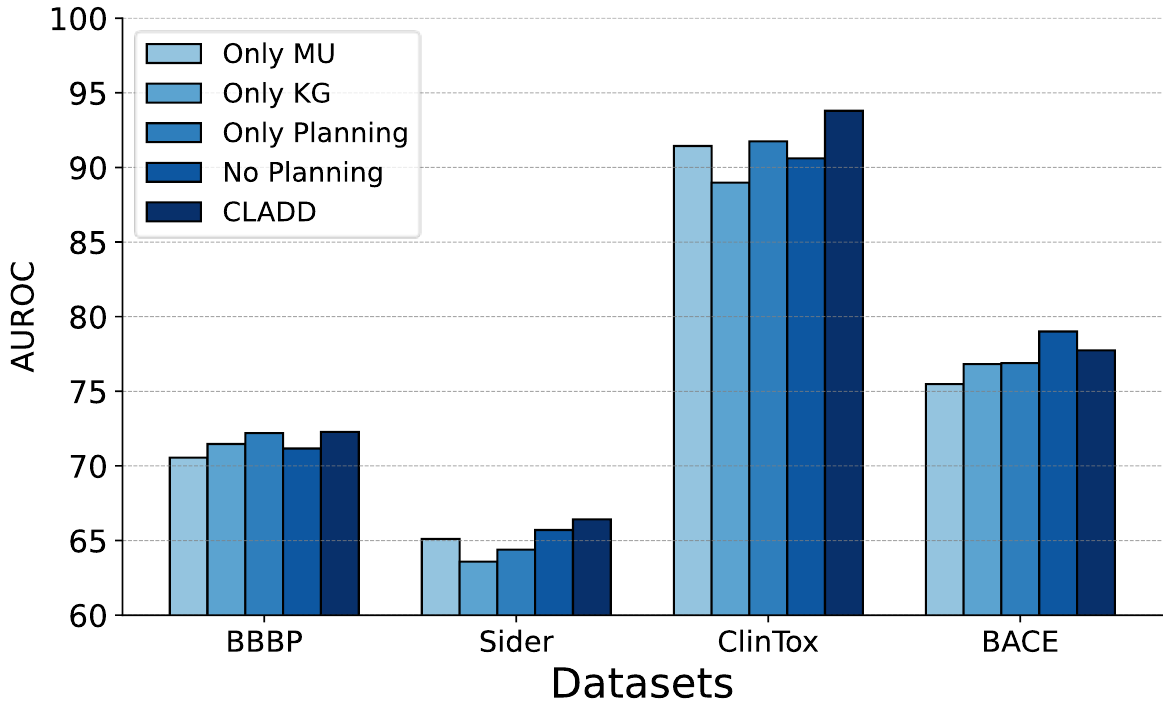}
        \vspace{-2ex}
        \caption{Ablation studies in the property-specific molecular captioning task.}
        \label{app fig: ablations captioning}
    \end{minipage}
\end{figure}

In Table \ref{app tab: ablations}, we conduct a model analysis by removing one component of the model at a time for the drug-target prediction task. We have the following observations:
\textbf{1)} By comparing ``Only Expert Annotation'' and ``Only Generated Caption'', we observe that relying solely on expert annotations yields significantly better performance. This highlights the critical importance of human-generated annotations over machine-generated captions. Still, their combination leads to the best overall performance.
\textbf{2)} Among the three agents—DrugRel Agent, BioRel Agent, and MU Agent—we could not determine a clear superiority in their relative importance, as it was task-dependent (Activation or Inhibition). 
\textbf{3)} Overall, we observe a decline in performance when any single component of \proposed~is removed, emphasizing the significance of each module.

We perform additional ablation studies in the property-specific molecule captioning task in Figure \ref{app fig: ablations captioning}.
Similarly, we observe that including all components (i.e., \proposed) leads to the best performance except for the BACE dataset.
Our analysis showed that this is because, as illustrated in Figure \ref{app fig: data planning}, the BACE dataset contains minimal relevant information in both the annotation database and the knowledge graph. 
Consequently, the model derives minimal benefit from external knowledge, highlighting the critical role of having relevant external information to boost performance.

\subsection{LLM-Agnostic Nature of CLADD}
\label{app: more llms}

Due to the expensive API costs, we mainly report the results using GPT-4o mini in the main manuscript to validate the proposed framework.
In this section, we performed additional experiments replacing it with different LLMs, including Llama3.3-70b and DeepSeek-V3. 
As shown in Table~\ref{tab: more llms}, the proposed framework (+ \proposed) consistently improves each individual LLM, showcasing its LLM-agnostic advantage.

\begin{table}[h!]
    \centering
    \caption{Performance of CLADD with each agent replaced by a different closed-source and open-source LLM (drug-target prediction task).}
    \resizebox{0.5 \linewidth}{!}{
    \begin{tabular}{lcccccccc}
    \toprule
    & & \multicolumn{2}{c}{\textbf{Overlap}} & & \multicolumn{2}{c}{\textbf{No overlap}} \\
    \cmidrule{3-4} \cmidrule{6-7}
    & & \textbf{Activate} & \textbf{Inhibit} & & \textbf{Activate} & \textbf{Inhibit}\\ \midrule
    GPT-4o mini & & 1.15 & 1.02 & & 1.13 & 0.87 \\
    + CLADD & & \textbf{3.04} & \textbf{4.83} & & \textbf{2.67} & \textbf{3.24} \\ \midrule
    Llama-3.3-70B & & 0.84 & 0.94 & & 0.88 & 0.81 \\
    + CLADD & & \textbf{3.13} & \textbf{6.40} & & \textbf{2.73} & \textbf{4.14} \\ \midrule
    DeepSeek-V3 & & 1.91 & 1.46 & & 1.90 & 1.11\\
    + CLADD & & \textbf{3.60} & \textbf{7.75} & & \textbf{3.15} & \textbf{5.01}\\ 
    \bottomrule
    \end{tabular}}
    \label{tab: more llms}
\end{table}

\subsection{Additional External Knowledge Analysis}
\label{app: additional external knowledge analysis}

\begin{figure}[h]
    \centering
    \begin{minipage}{0.45\linewidth}
    \centering
    \caption{Performance analysis on retrieval errors.}
    \resizebox{\linewidth}{!}{
    \begin{tabular}{lcccccccc}
    \toprule
    & & \multicolumn{2}{c}{\textbf{Overlap}} & & \multicolumn{2}{c}{\textbf{No overlap}} \\
    \cmidrule{3-4} \cmidrule{6-7}
    & & \textbf{Activate} & \textbf{Inhibit} & & \textbf{Activate} & \textbf{Inhibit}\\ \midrule
    GPT-4o mini & & 1.15 & 1.02 & & 1.13 & 0.87 \\
    Random Anchor Drug in KG & & 2.49 & 4.46 & & 1.86 & 2.31 \\
    Random Annotations in DB & & 2.62 & 4.08 & & 2.51 & 2.85 \\
    \midrule
    CLADD & & \textbf{3.04} & \textbf{4.83} & & \textbf{2.67} & \textbf{3.24} \\
    \bottomrule
    \end{tabular}}
    \label{tab: retrieval errors}
    \end{minipage}
    \hspace{1ex}
    \begin{minipage}{0.5\linewidth}
        \centering
        \includegraphics[width=0.95\linewidth]{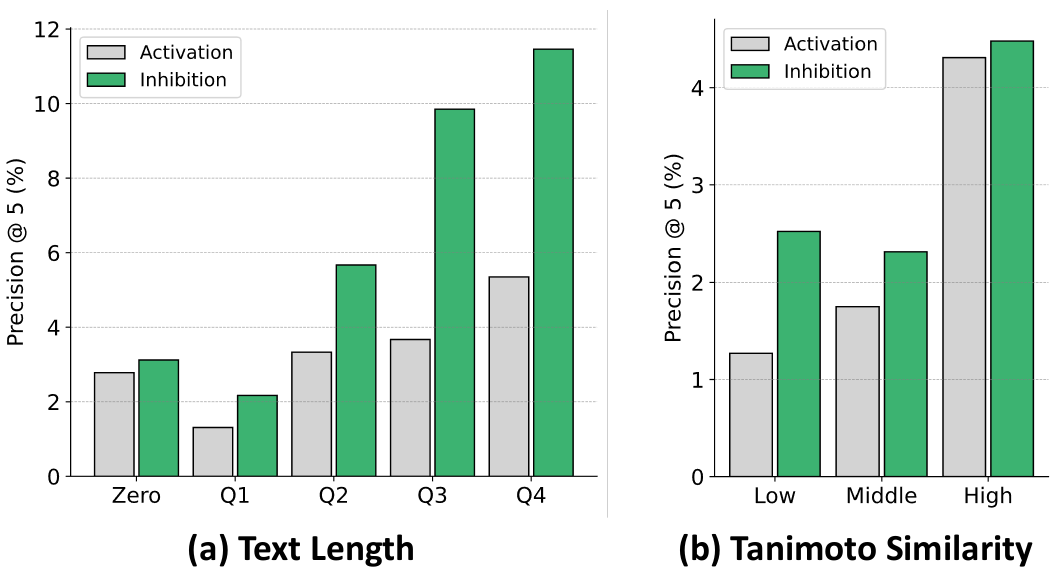}
        \caption{Performance as a function of (a) the length of the text retrieved from the annotation database and (b) the Tanimoto similarity between the anchor molecule and the knowledge graph.}
        \label{app fig: ret2perf}
    \end{minipage}
\end{figure}

In Table \ref{tab: retrieval errors}, we analyze how the retrieval accuracy affects the model performance. 
To do so,  we investigated two settings: one where the anchor drug selection in the knowledge graph is done randomly, and another where annotations are randomly sampled from the annotation database. 
As expected, we observe that the performance of both these models is significantly lower compared to the original model.
We also observe that there is still a significant performance gap when compared to GPT-4o mini. 
This is expected, as our model still includes a planning team that ensures that the anchor drug and annotations are only used when they are relevant to the query molecule and task.

Moreover, we further investigate how the quality of retrieved knowledge affects the model performance.
Firstly, we analyzed how performance changes as a function of the length of the annotation retrieved from the annotation database. 
In Figure \ref{app fig: ret2perf}(a), ``Zero" indicates that no annotation is available in the annotation database, while Q1, Q2, Q3, and Q4 represent the quartiles of the retrieved annotation length.
We highlight two interesting trends: (1) in general, performance increases with the annotation length, which is in line with the intuition that longer annotations include more relevant information, and (2) on average, “no annotation” leads to better results than the shortest annotations, which could indicate that the shortest annotations are often not informative enough to boost performance. However, for all groups except the shortest annotations, the additional information provides a proportional improvement.

Secondly, we analyzed how performance changes as a function of the similarity between the query molecule and the anchor molecule in the knowledge graph. 
In Figure \ref{app fig: ret2perf} (b), Molecules with a Tanimoto similarity of 1 are excluded from the evaluation. ``High": Tanimoto similarity between 0.7$\sim$1.0, ``Middle": Tanimoto similarity between 0.3$\sim$0.7, ``Low": Tanimoto similarity between 0.0$\sim$0.3.
Here, we found a very positive correlation, which is in line with the intuition that a higher similarity provides more relevant contextual information.

\begin{figure}[h]
    \centering
    \includegraphics[width=0.95\linewidth]{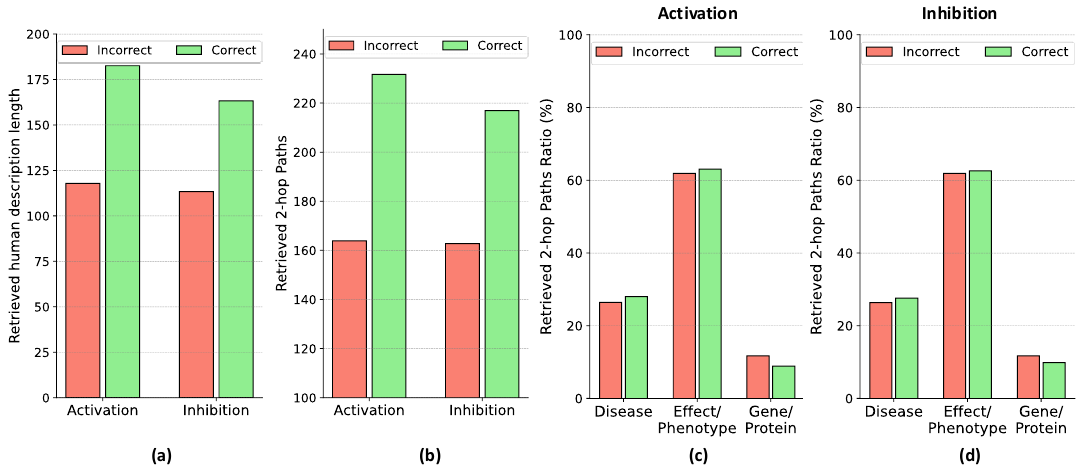}
    \caption{External knowledge analysis results. (a) The average length of retrieved human descriptions, (b) the average number of retrieved 2-hop paths in the knowledge graph, and (c-d) the proportion of entity types in 2-hop paths for correct and incorrect cases.}
    \label{app fig: model analysis}
\end{figure}

In Figure \ref{app fig: model analysis}, we analyze how external knowledge is used during the decision-making process for the drug-target prediction task.
We have the following observations:
\textbf{1)} As shown in Figures \ref{app fig: model analysis}(a) and \ref{app fig: model analysis}(b), the average length of human descriptions is considerably longer in the ``Correct" case, and the number of retrieved 2-hop paths is notably higher in the ``Correct" case. 
This highlights the importance of having external information that is both high quality and abundant.
\textbf{2)} On the other hand, although we anticipated a higher proportion of 2-hop paths containing Gene/Protein entities in the ``Correct" case, no significant difference was observed between the ``Correct" and ``Incorrect" cases in Figures \ref{app fig: model analysis}(c) and \ref{app fig: model analysis}(d).
From these results, we argue that \proposed's performance is not solely reliant on retrieving external information that is directly linked to the correct answer, given that external information can be further processed and contextualized by the agents, integrating different sources of evidence and internal knowledge.

\begin{figure}[h]
    \centering
    \includegraphics[width=0.8\linewidth]{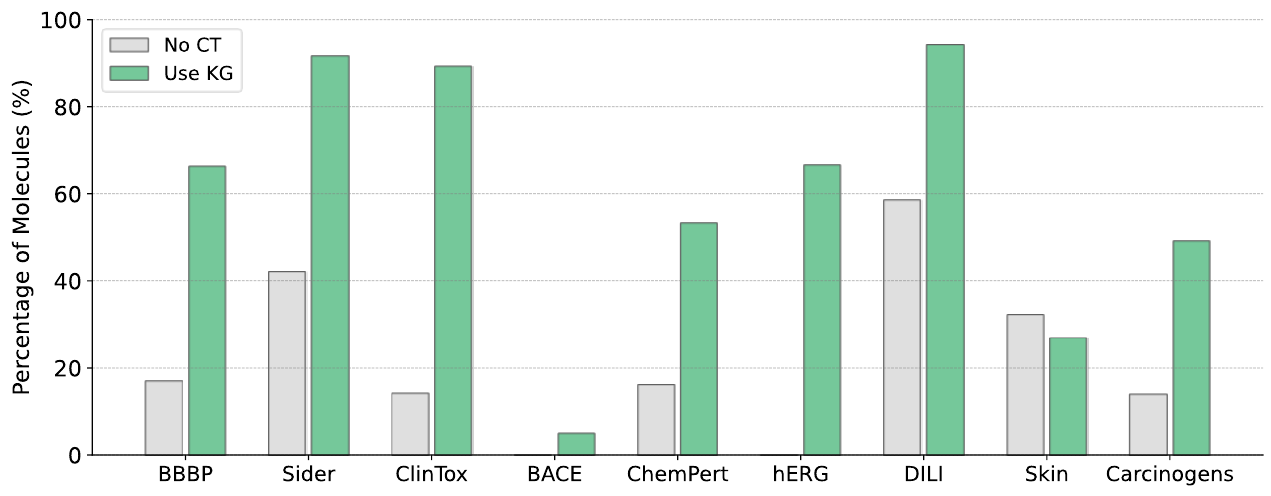}
    \caption{Planning team decision analysis based on different datasets. ``No CT" signifies that the planning team has decided not to utilize the captioning tool, while ``Use KG" indicates that the planning team intends to involve the Knowledge Graph Team.}
    \label{app fig: data planning}
\end{figure}

In Figure \ref{app fig: data planning}, we examine how the Planning Team determines the use of the captioning tool and collaborates with the Knowledge Graph Team based on the datasets.
We observed that, in most cases, the KG was used for more than 50\% of the query molecules, with the BACE and Skin Reaction datasets as significant exceptions. 
Furthermore, we observed that the BACE and hERG datasets lacked corresponding annotations for all query molecules.

\clearpage
\subsection{Additional Case Studies}
\label{app: additional case studies}
\begin{figure}[h]
    \centering
    \includegraphics[width=0.99\linewidth]{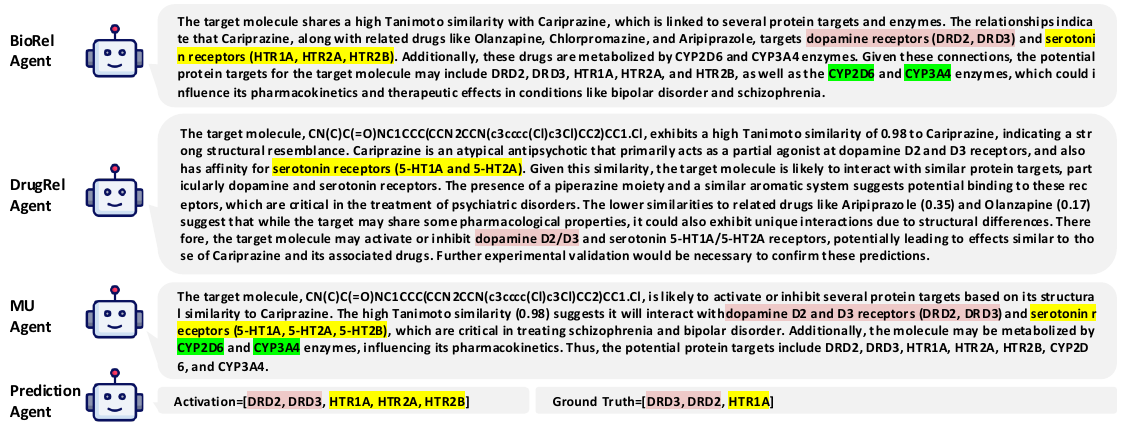}
    \caption{Additional case studies. Red represents dopamine-related proteins, yellow represents serotonin-related proteins, and green represents Cytochrome P450-related enzymes.}
    \label{app fig: qualitative}
\end{figure}

In this section, we provide additional details on case studies to analyze the behavior \proposed.
In Figure \ref{app fig: qualitative}, we observe that all three agents consistently predict dopamine-related and serotonin-related proteins as targets.
Based on the reports, Prediction Agent prioritizes these proteins over Cytochrome P450-related enzymes in the prediction.
Thus, we argue that our system can efficiently prioritize relevant information based on consensus, functioning similarly to a majority voting system.

\begin{figure}[h]
    \centering
    \includegraphics[width=0.99\linewidth]{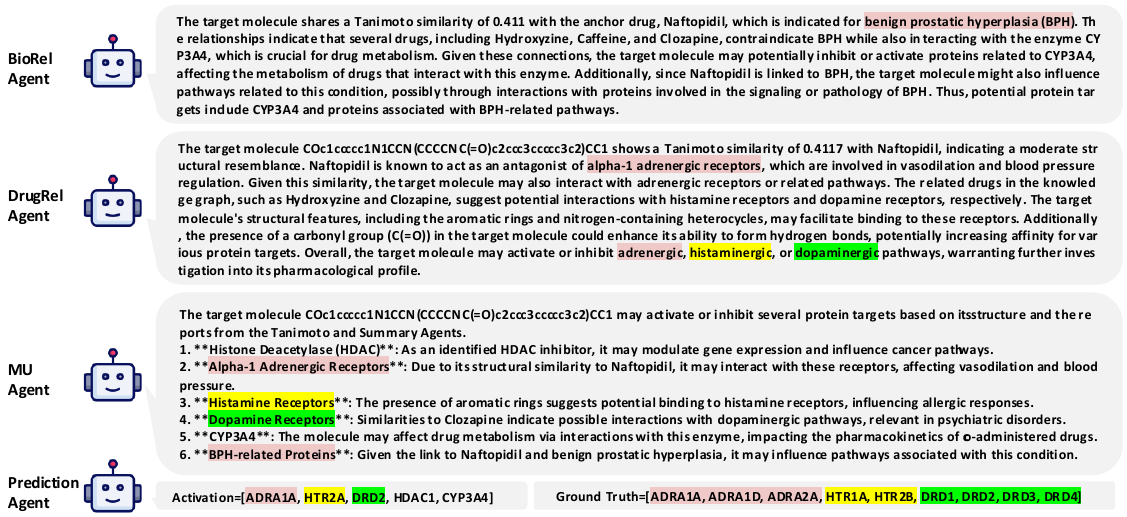}
    \caption{Full version of Figure \ref{fig: case studies}.}
    \label{app fig: qualitative full}
\end{figure}


\clearpage
\section{Agent Templates}
\label{app: agent templates}
In this section, we provide the templates for each agent used in Section \ref{sec: Methodology}.
We follow the previous work for designing the system prompt \cite{li2024empowering}.

\begin{table}[h!]
    \centering
    \caption{Prompts for Molecule Annotation Planner (Section \ref{sec: Planning Team}).}
    \resizebox{0.7\linewidth}{!}{
    \begin{tabular}{l}
    \toprule
\textbf{Prompt: } You are now working as an excellent expert in chemistry and drug discovery.\\
Your task is to determine whether the provided description is enough for analyzing \\the structure of the molecule. \\
\\
Are you ready?\\
\\
Description: \texttt{\textcolor{blue}{\{Retrieved Human Description\}}}\\
\\
You should answer in the following format:\\
\\
Answer = YES or NO\\
REASON = YOUR REASON HERE\\
\\
THERE SHOULD BE NO OTHER CONTENT INCLUDED IN YOUR RESPONSE.\\
    \bottomrule
    \end{tabular}}
    \label{app tab: prompt caption evaluator}
\end{table}
\begin{table}[h!]
    \centering
    \caption{Prompts for Knowledge Graph Planner (Section \ref{sec: Planning Team}).}
    \resizebox{0.7\linewidth}{!}{
    \begin{tabular}{l}
    \toprule
\textbf{Prompt: } You are now working as an excellent expert in chemistry and drug discovery.\\
Your task is to decide whether to utilize the knowledge graph structure by evaluating the structural \\ similarity between the target molecule and the anchor drug within the knowledge graph.\\
If the target molecule and the anchor drug show high similarity, the knowledge graph should be \\ leveraged to extract relevant information.\\
\\
The Tanimoto similarity between the target molecule \texttt{\textcolor{blue}{\{SMILES\}}} and the anchor drug \\
\texttt{\textcolor{blue}{\{SMILES\}}} (\texttt{\textcolor{blue}{\{Drug Name\}}}) is \texttt{\textcolor{blue}{\{Tanimoto Similarity\}}}.\\
\\
You should answer in the following format:\\
\\
Answer = YES or NO\\
REASON = YOUR REASON HERE\\
\\
THERE SHOULD BE NO OTHER CONTENT INCLUDED IN YOUR RESPONSE.\\
    \bottomrule
    \end{tabular}}
    \label{app tab: prompt kg evaluator}
\end{table}
\begin{table}[h!]
    \centering
    \caption{Prompts for Biology Relation Agent (Section \ref{sec:KnowledgeGraphTeam}).}
    \resizebox{0.7\linewidth}{!}{
    \begin{tabular}{l}
    \toprule
\textbf{Prompt: } You are now working as an excellent expert in chemistry and drug discovery.\\
Your task is to predict \texttt{\textcolor{blue}{\{Task Description\}}} by analyzing the relationships between the anchor drug, \\which shares tanimoto similarity of \texttt{\textcolor{blue}{\{Tanimoto Similarity\}}} with the target molecule, \\and the most closely related drugs in the knowledge graph. \\
\\
You should explain the reasoning based on the intermediate nodes between the \\related drugs and the anchor drug, as well as the types of relationships they have.\\
\\
The two-hop relationships between the drugs will be provided in the following format:\\
(Drug A, relation, Entity, relation, Drug B), where the entity can be one of the following \\three types of entities: (gene/protein, effect/phenotype, disease)\\
\\
Are you ready?\\
\\
Target molecule: \texttt{\textcolor{blue}{\{SMILES\}}}\\
\\
Here are the two-hop relationships:\\
\texttt{\textcolor{blue}{\{Two-hop Paths\}}}\\
\\
DO NOT ANSWER IN THE PROVIDED FORMAT.\\
DO NOT WRITE MORE THAN 300 TOKENS.\\
THERE SHOULD BE NO OTHER CONTENT INCLUDED IN YOUR RESPONSE.\\
    \bottomrule
    \end{tabular}}
    \label{app tab: prompt bra}
\end{table}
\begin{table}[h!]
    \centering
    \caption{Prompts for Drug Relation Agent (Section \ref{sec:KnowledgeGraphTeam}).}
    \resizebox{0.7\linewidth}{!}{
    \begin{tabular}{l}
    \toprule
\textbf{Prompt: } You are now working as an excellent expert in chemistry and drug discovery.\\
\\
Your task is to \texttt{\textcolor{blue}{\{Task Description\}}} by analyzing its structural similarity to anchor drugs \\and related drugs, and provide an explanation grounded in its resemblance to these other drugs.\\
\\
Are you ready?\\
\\
The Tanimoto similarity between the target molecule \texttt{\textcolor{blue}{\{SMILES\}}} and the anchor drug \texttt{\textcolor{blue}{\{SMILES\}}} \\ (\texttt{\textcolor{blue}{\{Drug Name\}}} is \texttt{\textcolor{blue}{\{Tanimoto Similarity\}}}.\\
\\
The anchor drug \texttt{\textcolor{blue}{\{Drug Name\}}} is highly associated with the following molecules \\in the knowledge graph: \texttt{\textcolor{blue}{\{Reference Drugs\}}}.\\
\\
The Tanimoto similarities between the target molecule \texttt{\textcolor{blue}{\{SMILES\}}} and the related drugs \\in the knowledge graph are \texttt{\textcolor{blue}{\{Tanimoto Similarity\}}}.\\
\\
DO NOT WRITE MORE THAN 300 TOKENS.\\
THERE SHOULD BE NO OTHER CONTENT INCLUDED IN YOUR RESPONSE.\\
    \bottomrule
    \end{tabular}}
    \label{app tab: prompt dra}
\end{table}
\begin{table}[h!]
    \centering
    \caption{Prompts for Molecule Understanding Agent (Section \ref{sec: MolecualrUnderstandingTeam}).}
    \resizebox{0.7\linewidth}{!}{
    \begin{tabular}{l}
    \toprule
\textbf{Prompt: } You are now working as an excellent expert in chemistry and drug discovery.\\
\\
Your task is to predict \texttt{\textcolor{blue}{\{Task Description\}}} by using the SMILES representation \\and description of a molecule, and explain the reasoning based on its description.\\
\\
You can also consider the report from other agents involved in drug discovery:\\
- Drug Relation Agent: Evaluates the structural similarity between the target molecule and related molecules.\\
- Biology Relation Agent: Examines the biological relationships among the related molecules.\\
\\
Are you ready?\\

SMILES: \texttt{\textcolor{blue}{\{SMILES\}}}\\
Description: \texttt{\textcolor{blue}{\{Caption\}}}\\
\\
Below is the report from other agents.\\

Drug Relation Agent:\\
\texttt{\textcolor{blue}{\{Report from Drug Relation Agent\}}}\\
\\
Biology Relation Agent:\\
\texttt{\textcolor{blue}{\{Report from Biology Relation Agent\}}}\\
\\
DO NOT WRITE MORE THAN 300 TOKENS.\\
THERE SHOULD BE NO OTHER CONTENT INCLUDED IN YOUR RESPONSE.\\
    \bottomrule
    \end{tabular}}
    \label{app tab: prompt tma}
\end{table}
\begin{table}[h!]
    \centering
    \caption{Prompts for Prediction Agent (Section \ref{sec:PredictionAgent}).}
    \resizebox{0.7\linewidth}{!}{
    \begin{tabular}{l}
    \toprule
\textbf{Prompt: } You are now working as an excellent expert in chemistry and drug discovery.\\
\\
Your task is to predict \texttt{\textcolor{blue}{\{Task Description\}}} \texttt{\textcolor{blue}{\{SMILES\}}}.\\
\\
Your reasoning should be based on reports from various agents involved in drug discovery:\\
- Molecule Understanding Agent: Focuses on analyzing the structure of the target molecule.\\
- Drug Relation Agent: Evaluates the structural similarity between the target molecule and related molecules.\\
- Biology Relation Agent: Examines the biological relationships among the related molecules.\\
\\
Below is the report from each agent.\\
\\
Molecule Understanding Agent:\\
\texttt{\textcolor{blue}{\{Report from Molecule Understanding Agent\}}}\\
\\
Drug Relation Agent:\\
\texttt{\textcolor{blue}{\{Report from Drug Relation Agent\}}}\\
\\
Biology Relation Agent:\\
\texttt{\textcolor{blue}{\{Report from Biology Relation Agent\}}}\\
\\
Based on the reports, \texttt{\textcolor{blue}{\{Task Description and Answering Format\}}}\\
\\
THERE SHOULD BE NO OTHER CONTENT INCLUDED IN YOUR RESPONSE.\\
    \bottomrule
    \end{tabular}}
    \label{app tab: prompt task agent}
\end{table}


\end{document}